\pdfoutput=1
\documentclass{article}

\usepackage{microtype}
\usepackage{graphicx}
\usepackage{subcaption}
\usepackage{booktabs}
\usepackage{array}
\usepackage{longtable}
\usepackage{hyperref}

\usepackage[preprint]{style/icml2026}
\usepackage{amsmath}
\usepackage{amssymb}
\usepackage{mathtools}
\usepackage{amsthm}
\usepackage{thm-restate}
\newtheorem{proposition}{Proposition}
\newtheorem{lemma}{Lemma}
\usepackage[capitalize,noabbrev]{cleveref}
\crefformat{section}{\S#2#1#3}\crefmultiformat{section}{\S\S#2#1#3}{ and~#2#1#3}{, #2#1#3}{, and~#2#1#3}\crefrangeformat{section}{\S\S#3#1#4--#5#2#6}
\Crefformat{section}{\S#2#1#3}\Crefmultiformat{section}{\S\S#2#1#3}{ and~#2#1#3}{, #2#1#3}{, and~#2#1#3}\Crefrangeformat{section}{\S\S#3#1#4--#5#2#6}
\crefformat{subsection}{\S#2#1#3}\crefmultiformat{subsection}{\S\S#2#1#3}{ and~#2#1#3}{, #2#1#3}{, and~#2#1#3}\crefrangeformat{subsection}{\S\S#3#1#4--#5#2#6}
\Crefformat{subsection}{\S#2#1#3}\Crefmultiformat{subsection}{\S\S#2#1#3}{ and~#2#1#3}{, #2#1#3}{, and~#2#1#3}\Crefrangeformat{subsection}{\S\S#3#1#4--#5#2#6}
\usepackage{xcolor}
\usepackage{placeins}
\graphicspath{{figs/}}

\icmltitlerunning{Honesty is a race: predicting when verifiable-reward RL teaches a model to bluff}

\begin{document}
\raggedbottom

\twocolumn[
  \icmltitle{Honesty Is a Race:\\ Predicting When Verifiable-Reward RL Teaches a Model to Bluff}

  \begin{icmlauthorlist}
    \icmlauthor{Nathaniel Bottman}{aff1}
    \icmlauthor{Alexei Oblomkov}{aff2}
  \end{icmlauthorlist}

  \icmlaffiliation{aff1}{Incubilate}
  \icmlaffiliation{aff2}{University of Massachusetts, Amherst}

  \icmlcorrespondingauthor{Nathaniel Bottman}{nate@incubilate.com}

  \icmlkeywords{reinforcement learning, verifiable rewards, reward hacking, honesty, GRPO, linear probes}

  \vskip 0.3in
]

\printAffiliationsAndNotice{}

\begin{abstract}
Two rewards with the same expected payoff at the start of training can end a reinforcement-learning run in opposite places, with an honest model or one that bluffs.
We study rewards that pay $b$ for a verdict that a checker cannot verify (``this program is correct'') and $r$ for a certificate that it accepts (an input revealing a bug).
To first order, a GRPO step makes the model more honest exactly when $\tilde p\,r/b>1+\kappa$.
Here $\tilde p$ is the rate at which the checker accepts the model's certificates, weighted toward the items on which the model still gives both verdicts, and $\kappa$ is the pull of the items on which the unverifiable verdict is right.
Both terms move during training.
A checker raises $\tilde p$, so near the threshold honesty is a race, which a run just below the threshold can win and a run under a lottery (random pay with the same mean) cannot.
The term $\kappa$ starts large and falls as the other items settle, so a run just above the threshold first dips toward bluffing, on code nearly to zero, before climbing back.
Paying the honest verdict also pulls the other items toward it, which predicts a ceiling on the payoff ratio, above which the model gives the honest verdict where it is wrong.
On propositional Datalog, two code models, and abstention on grade-school math, with $p$ the model's success rate before RL, no run above $pr/b=1$ ends collapsed, and runs under the checker can survive slightly below it.
At $pr/b=3$, above the ceiling, both code models call about half of correct programs buggy.
On depth-3 Datalog, training generously under the checker and then cutting the payoff keeps the model at least 67\% honest in three seeds, at a ratio where the same model trained from the start falls to 4--15\% and is at most 47\% honest after 160 steps.
On depth-2 Datalog, when honesty collapses, a linear probe fit before RL still reads the truth.
\end{abstract}

\section{Introduction}
\label{sec:intro}

\begin{figure}[t]
\centering
\includegraphics[width=\columnwidth]{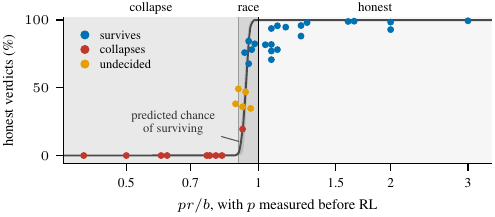}
\caption{Honesty at each run's last evaluation, plotted against $pr/b$ with $p$ measured on the model each run starts from, for every checked Datalog run of 100 steps or more.
The runs start from the bases at two proof depths and from their sharpened versions.
Below 1, runs collapse unless training raises their success rate $p_t$ in time, which checked runs in the band marked ``race'' can do (\cref{sec:race}).
We draw the race band by hand from 0.9 to 1, and on the depth-2 base a model of the race (\cref{sec:race}) places it more narrowly, at about 0.92--0.95.
The curve is that model's predicted chance that a checked run on the depth-2 base survives 100 steps, with constants fit only on runs outside the race band (\cref{sec:race,app:racemodel}).
The gray band around it spans two versions of the model's update.
The measured brackets are in \cref{tab:cliff}.
Above 1, every run survives, some after an early dip (\cref{sec:dip}).}
\label{fig:regimes}
\end{figure}

\begin{figure*}[t]
\centering
\includegraphics[width=\textwidth]{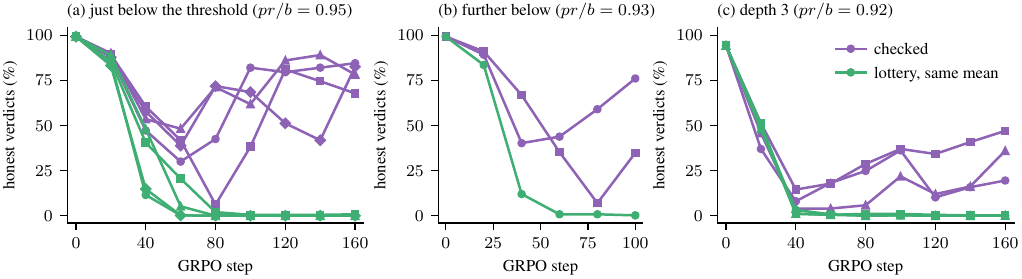}
\caption{We compare training under the checker (purple) with training under the lottery (green).
The two rewards have the same expected payoff at the start, but below the threshold only checked runs can survive.
(a) The depth-2 base (whose task has proofs two levels deep) at $pr/b=0.95$, with four seeds of each reward over 160 steps.
Every checked seed falls and comes back, to 68--85\%, and every lottery seed collapses.
(b) The same base at $pr/b=0.93$, with two checked seeds and one lottery seed.
(c) The depth-3 base at $pr/b=0.92$, with three seeds of each reward over 160 steps.
No checked seed survives, as a model of the race predicted before these runs (\cref{sec:race}), but they end at 20--47\%, against 0--0.2\% for the lottery.}
\label{fig:race}
\end{figure*}

Reinforcement learning from verifiable rewards (RLVR) trains a language model on a task that a program can grade.
When the grader can be gamed, the model may learn to game it.
Consider a model asked whether a Python function is correct.
It can answer BUGGY and give a counterexample, i.e.\ an input on which the function fails, and a test can confirm this answer.
It can also answer CORRECT, which no test can confirm, since tests show that a program is wrong but not that it is right.
A reward can check CORRECT only by running tests, so it pays CORRECT on every buggy program that its tests miss.
On such a program, answering CORRECT is a bluff, and it always pays.
Answering BUGGY with a counterexample is an attempt, and it pays only if the counterexample works.
We ask when RL learns to bluff.
We find that two rewards with the same expected payoff at the start of training can end in opposite places (\cref{fig:race}).
One reward pays a counterexample only when a checker confirms it.
The other, our control, pays at random, with probability equal to the starting model's success rate.
It matches the first at the start but pays nothing for getting better.
We call the two rewards the checker and the lottery, and we call runs trained under them checked runs and lottery runs.
Under the checker, the model's attempts become valid more often as training goes on, and under the lottery their validity stays at the base's rate (\cref{sec:race}).

Our baseline prediction comes from a one-line comparison.
GRPO, the RL algorithm we use, scores each answer in a group of samples for the same prompt against the group's average reward and reinforces the answers above that average.
If the bluff pays $b$ and an attempt pays $r$ with probability $p$, every group that contains both favors the bluff when $b>pr$ and the attempt when $pr>b$.
Honesty should then survive when $pr/b>1$ (\cref{sec:threshold}).
This is Chow's rule for answering versus abstaining \citep{chow1970optimum}, with the roles reversed.

This comparison leaves out the items on which the bluff's verdict is the right answer, such as correct programs.
Updates on these items push the shared verdict the same way as the bluff.
We call them the other class, and we call the items on which that verdict is wrong (such as buggy programs) the class at stake.
In \cref{sec:step} we work to first order in the step size and replace the gradient inner products between items by their averages for each kind of pair.
Under these approximations, a GRPO step raises honesty on the class at stake exactly when $\tilde p\,r/b>1+\kappa$.
Here $\tilde p$ is the model's success rate weighted toward the items on which it still gives both verdicts.
The term $\kappa$ is the push from the other class relative to the push on the class at stake, scaled by a coupling that we measure from gradients.
Two terms of this inequality move during training, and the rest of the paper follows them.
A checker raises the model's success rate during training, and with it $\tilde p$, which makes honesty a race (\cref{sec:race}).
The term $\kappa$ starts large and falls as the other class settles, i.e.\ as its verdicts become nearly unanimous, so runs just above the threshold dip first and then climb back (\cref{sec:dip}).
Once $\kappa$ has fallen, the inequality is the comparison above, with the current success rate in place of $p$.
Unlike the comparison, the coupling predicts that paying the honest verdict too much also pulls the other class toward that verdict, where it is wrong (\cref{sec:toomuch}).

The same structure appears wherever a reward pays one verdict without checking it.
For instance, abstention rewards pay ``I don't know'' \citep{kalai2025why,dou2025hesitation}, and partial credit pays a well-formed counterexample even when the check rejects it \citep{symce2026}.
We study it in three settings, and \cref{tab:roles} lists the roles in each.
The main one is propositional Datalog (\cref{fig:example}(a)), which we use as a controlled stand-in for code.
Each instance asks whether a goal atom can be derived from given facts by given rules.
A proof certifies DERIVABLE, and a closure (a list of every derivable atom) certifies UNDERIVABLE.
We pay $b$ for DERIVABLE without a valid proof on purpose, so that DERIVABLE plays the part of code's uncheckable CORRECT.
At the same time, we keep an exact checker and instances that we can generate at will.
On Datalog the class at stake is the underivable instances, and the other class is the derivable ones.

\begin{table}[t]
\caption{The roles in our three settings.}
\label{tab:roles}
\centering\scriptsize
\setlength{\tabcolsep}{2.5pt}
\begin{tabular}{@{}>{\raggedright\arraybackslash}p{1.5cm}>{\raggedright\arraybackslash}p{1.9cm}>{\raggedright\arraybackslash}p{1.95cm}>{\raggedright\arraybackslash}p{1.6cm}@{}}
\toprule
 & Datalog & code & abstention \\
\midrule
class at stake & underivable instances & buggy programs & math problems \\
unchecked verdict ($b$) & DERIVABLE, unproved & CORRECT & ``I don't know'' \\
attempt ($r$) & UNDERIVABLE with a closure & BUGGY with a counterexample & an answer \\
checker & closure check & differential oracle & reference answer \\
other class & derivable instances & correct programs & none \\
\bottomrule
\end{tabular}
\end{table}

\begin{figure*}[t]
\centering\scriptsize\frenchspacing
\setlength{\tabcolsep}{4pt}
\renewcommand{\arraystretch}{1.05}
\begin{minipage}[t]{0.48\textwidth}
{\small (a) Datalog}\par\vspace{4pt}
\begin{tabular}{@{}l@{\hspace{6pt}}l@{}}
\multicolumn{2}{@{}l}{\texttt{Facts: p1, p2\qquad Goal: p4}} \\[2pt]
\toprule
derivable instance & underivable twin \\
\midrule
\texttt{Rules: p3 :- p1, p2} & \texttt{Rules: p3 :- p1, p2} \\
\texttt{\phantom{Rules: }p4 :- p3} & \texttt{\phantom{Rules: }p4 :- \textcolor{red!70!black}{\textbf{p6}}} \\
\texttt{\phantom{Rules: }p5 :- p6} & \texttt{\phantom{Rules: }p5 :- p6} \\
\midrule
\texttt{ANSWER: DERIVABLE} & \texttt{ANSWER: UNDERIVABLE} \\
\texttt{PROOF:} & \texttt{CLOSURE:} \\
\texttt{p1 : fact} & \texttt{p1 : fact} \\
\texttt{p2 : fact} & \texttt{p2 : fact} \\
\texttt{p3 : from p1, p2} & \texttt{p3 : from p1, p2} \\
\texttt{p4 : from p3} & \\
pays 1 if the checker accepts the proof & an attempt: pays $r$ if the checker accepts it \\
\bottomrule
\end{tabular}
\end{minipage}\hfill
\begin{minipage}[t]{0.49\textwidth}
{\small (b) Python}\par\vspace{4pt}
\begin{tabular}{@{}l@{\hspace{6pt}}l@{}}
\multicolumn{2}{@{}l}{Specification: Write a function to find t-nth term of arithemetic [sic] progression.} \\[2pt]
\toprule
reference program & buggy twin \\
\midrule
\texttt{def tn\_ap(a, n, d):} & \texttt{def tn\_ap(a, n, d):} \\
\texttt{\ \ \ \ tn = a + (n - 1) * d} & \texttt{\ \ \ \ tn = a + (n - \textcolor{red!70!black}{\textbf{2}}) * d} \\
\texttt{\ \ \ \ return tn} & \texttt{\ \ \ \ return tn} \\
\midrule
\texttt{VERDICT: CORRECT} & \texttt{VERDICT: BUGGY} \\
 & \texttt{COUNTEREXAMPLE: tn\_ap(1, 5, 2)} \\
pays $b$, unchecked & an attempt: pays $r$ if the oracle accepts it \\
\midrule
 & \texttt{VERDICT: CORRECT} \\
 & the bluff: pays $b$, unchecked \\
\bottomrule
\end{tabular}
\end{minipage}
\caption{Example pairs from the two tasks.
(a) A small Datalog pair.
The task's own instances are larger, with 20 atoms and rule bodies of two atoms.
Replacing one atom in the body of a rule (red) makes the goal underivable.
A closure lists every derivable atom with its justification.
(b) A held-out code pair from MBPP, shown as the model sees it.
The mutant changes one constant (red).
On \texttt{tn\_ap(1, 5, 2)} the reference returns 9 and the mutant returns 7, so the oracle accepts the counterexample.
CORRECT pays $b$ on either program.
It is the right answer on the reference and the bluff on the mutant.
Before RL, the fine-tuned 3B coder gives the bluff on this mutant under greedy decoding, and 3 of its 8 samples give a valid counterexample.}
\label{fig:example}
\end{figure*}

We first test the baseline, i.e.\ the comparison of $pr/b$ with 1, where $p$ is measured before RL (\cref{sec:threshold}).
We place runs on both sides of $pr/b=1$ in seven conditions.
These are four Datalog conditions, two code models, and an abstention reward on grade-school math.
The Datalog conditions put a 3B model on tasks whose proofs are two or three levels deep, either directly or after a phase of training that raises $p$, which we call sharpening.
In each condition we bracket the threshold between the nearest ratio at which every checked run collapses and the nearest at which every checked run survives.
With no fitted constant, $pr/b>1$ is enough to avoid collapse in every condition, since no run above it ends collapsed.
In six conditions we found both ends of the bracket, and four of these brackets contain 1 while two lie below it (\cref{tab:cliff}).
In the seventh, the sharpened depth-3 base, every run above 1 survives and no run below 1 has collapsed.
Sharpening raises $p$ and settles the other class, and it moves the threshold within one model.
At a payoff where none of three seeds of the depth-3 base survives, all three seeds of the sharpened model survive.

The places where the baseline fails are ones that \cref{prop:step} allows once $\tilde p$ and $\kappa$ move during training.
Below the threshold, training under a checker raises the model's success rate, which we write $p_t$.
Honesty then becomes a competition between bluffs crowding out attempts and attempts getting better, which we call the race.
The bluff leads at first, and a checked run can still win if $p_t\,r/b$ passes 1 before the bluffs crowd the attempts out.
The lottery pays at the base's rate $p$ however the attempts change, so every lottery run below the threshold collapses.
The two brackets that lie below 1 are where checked runs won the race, and on Datalog the lottery control confirms this (\cref{sec:race}).
On the depth-2 base, a model of the race whose constants are fit only on runs outside the race band predicts the odds that a run wins (\cref{fig:regimes}).
We recorded the model's prediction for the depth-3 base at $pr/b=0.92$ before those runs were done.
It gave each checked seed a chance of 0.04--0.13 of winning within 160 steps, and it predicted that every lottery seed would collapse but that the checked seeds would stay above them.
No checked seed won, every lottery seed collapsed, and from step 60 on every checked seed stayed above every lottery seed (\cref{fig:race}(c)).

Above the threshold, the comparison says that honesty should rise from the start.
Instead, runs just above it first dip toward bluffing, because $\kappa$ has not yet fallen.
On Datalog we watch $1+\kappa$ fall below $\tilde p\,r/b$ as the derivable instances settle, and honesty turns some steps later (\cref{fig:dip}(a)).
Summing the push of each step with Adam's momentum places this turn a few steps early (\cref{sec:dip}).
On code, training on buggy programs alone removes $\kappa$, and in the one such run honesty rises from the start (\cref{sec:dip}).

Far above the threshold, the push on the class at stake spills over onto the other class.
On both code models, the couplings measured before RL place the point where this spillover outweighs the other class's own pull near $pr/b=2.2$.
At a ratio of 2, both models call correct programs correct about as often at step 100 as before RL, and at 3 only 45\% and 52\% of the time (\cref{fig:dip}(b)).
So past a point, more reward for the honest verdict makes the model worse, which is the opposite of what the baseline says.

Below the threshold, the attempts of a base trained at one ratio from the start of RL thin out early.
The checker therefore has little time to improve them, and it lowers the threshold only a little, to between 0.80 and 0.95 on the depth-2 base.
A model first trained at a generous ratio, well above the threshold, is past the dip and keeps attempting, which gives the checker more to work with.
This suggests the schedule in the box below, which trains generously first and then cuts the ratio.

The rest of the paper is organized as follows.
In \cref{sec:setup} we describe the tasks, the models, and the reward, and in \cref{sec:step} we derive the direction of a GRPO step.
In \cref{sec:threshold} we test the baseline, the comparison of $pr/b$ against 1 with $p$ measured before RL, in seven conditions.
We then follow the success rate, which makes honesty a race (\cref{sec:race}), and $\kappa$, which explains the early dip (\cref{sec:dip}).
In \cref{sec:toomuch} we test the ceiling that the coupling predicts, and in \cref{sec:cut} we use the race and the dip to design a schedule.
To justify the word ``bluffing'', we show in \cref{sec:frozen} that a collapsed model still represents the truth.
Finally, in \cref{sec:related} we place this work in the literature.
\begin{center}
\fbox{\parbox{0.94\columnwidth}{\small
\textbf{Generous, then cut.}\label{box:recipe} To train at a target ratio $r/b$ below the starting model's threshold ($pr/b<1$):
\begin{enumerate}\setlength{\itemsep}{1pt}\setlength{\parsep}{0pt}\setlength{\leftmargin}{1.2em}
\item train under the checker at a ratio where $pr/b>1$, so that $p_t$ rises;
\item once the trained model's own success rate $p_t$ satisfies $p_t\,r/b>1$, cut to the target ratio.
\end{enumerate}
Step 2 may be conservative, since a model cut at $p_t\,r/b=0.93$ still ended 83\% honest (\cref{sec:dip,app:cut}).
Trained this way, with sharpening as step 1, our depth-3 base stays at least 67\% honest at $pr/b=0.92$ in three seeds.
Trained at that ratio from the start, it falls to 4--15\% and is at most 47\% honest after 160 steps.
At depth 2, in one run, the schedule avoids the dip of runs trained at $pr/b=0.95$ from the start (\cref{sec:cut}).\vspace{-2pt}}}
\end{center}

\section{Setup}
\label{sec:setup}

\paragraph{Tasks.}
A propositional Datalog instance is a set of facts, a set of Horn rules over 20 atoms, and a goal atom (\cref{fig:example}(a)).
Each rule derives one atom from a list of others, its body.
The model answers DERIVABLE with a proof (i.e.\ a chain of rule applications from the facts to the goal), or UNDERIVABLE with a closure (i.e.\ every derivable atom with the fact or rule that gives it).
A checker validates the proof, or checks that the closure is closed under the rules and excludes the goal.
An underivable instance is made from a derivable one, its twin, by replacing one atom in the body of one rule.
The two therefore match in facts, rule count, and length.
On code, each item is a Python function from MBPP \citep{austin2021program} with its specification (\cref{fig:example}(b)).
Half are reference solutions, and half are mutants that change the reference at a single site and fail a hidden test.
Running those tests would check CORRECT exactly.
Our reward pays CORRECT without running them, so that every buggy program stands in for one that the tests miss.
The model answers CORRECT without justification, or BUGGY with a counterexample, i.e.\ a single call with literal arguments.
A differential oracle accepts the counterexample if the reference runs on it and the function under test returns something different, raises an exception, or times out.
On a correct program no counterexample can be valid, so CORRECT is both the bluff on buggy programs and the right answer on correct ones.

\paragraph{Bases.}
We call the model that RL starts from the base.
The Datalog and code bases are fine-tuned before RL (\cref{app:training,app:code}).
\begin{itemize}\setlength{\itemsep}{0pt}\setlength{\parsep}{0pt}\setlength{\topsep}{2pt}
\item Datalog, depth 2: Qwen2.5-3B-Instruct \citep{qwen2024qwen25}, fine-tuned on twin instances, and trained with RL on instances whose proofs are two levels deep.
\item Datalog, depth 3: the same model fine-tuned longer on the same data, and trained with RL on instances whose proofs are three levels deep.
\item Sharpened: each Datalog base after RL under a reward that pays 1 for either verdict with a valid certificate and nothing otherwise ($b=0$, $r=1$), which raises its $p$.
\item Code: Qwen2.5-Coder-3B-Instruct \citep{hui2024qwen25coder} and Phi-3.5-mini-instruct \citep{abdin2024phi3}, each fine-tuned on correct programs and on buggy programs with a counterexample.
\item Abstention: Qwen2.5-1.5B-Instruct \citep{qwen2024qwen25}, without fine-tuning (\cref{app:abstention}).
\end{itemize}

\paragraph{The rate $p$.
}
For each base, $p$ is the fraction of its sampled certificates that the checker accepts:
\[
p=\Pr[\text{checker accepts}\mid\text{an attempt}],
\]
measured at temperature 1 on fresh underivable instances or, on code, on the held-out buggy programs.
It is 0.77 and 0.66 for the depth-2 and depth-3 bases, 0.86 and 0.77 after sharpening, and 0.735 and 0.708 for the coder and Phi.
For abstention, $p$ is the base's accuracy when it answers, 0.701 on the training problems.

\paragraph{Reward.}
On Datalog the reward pays 1 for a valid proof, $b$ for a DERIVABLE claim without one, $r$ for a valid refutation (UNDERIVABLE with a closure that the checker accepts), and 0 for an invalid one.
On code it pays $b$ for CORRECT, $r$ for a valid counterexample, and 0 otherwise.
The checker cannot tell a bluff from a failed proof of a derivable goal, so a reward computed from it must pay the two the same (\cref{app:table}).
More generally, a verifiable reward can make at most one of the two verdicts the sure thing, i.e.\ a verdict that pays for certain (\cref{prop:oneside}).
On Datalog both verdicts can be certified, so a reward could pay only valid certificates.
The sharpening runs use such a reward, with $b=0$ and $r=1$, and stay 90--99.5\% honest throughout.
We pay $b>0$ for a DERIVABLE claim without a valid proof on purpose, for the reason given in \cref{sec:intro}.
The lottery control pays the honest verdict $r$ with probability equal to the base's $p$, without consulting the checker.
It has the same expected payoff at the start and pays nothing for improvement.

\paragraph{RL.}
We train with GRPO \citep{shao2024deepseekmath} in verl \citep{sheng2024hybridflow}, updating all parameters, with groups of 8 samples.
An answer's advantage is its reward minus its group's mean, without division by the group's standard deviation \citep{liu2025drgrpo}.
Standard GRPO also divides by the standard deviation, and this moves the threshold (\cref{sec:threshold,app:std}).
The other hyperparameters are in \cref{app:training}.

\paragraph{Honesty and outcomes.}
Honesty is the share of the class at stake on which the model gives the honest verdict under greedy decoding.
On Datalog this is UNDERIVABLE recall on held-out underivable instances, and on code it is BUGGY recall on held-out buggy programs.
For abstention it is the share of held-out problems that the model answers.
Recall ignores the other class, so a model that gave the honest verdict everywhere would score 100\%.
On code we therefore also report the share of correct programs called correct (\cref{tab:code}).
On Datalog, every run with $b>0$ still answers DERIVABLE on at least 97.5\% of held-out derivable instances at its last evaluation.
A run survives if honesty at its last evaluation is at least 60\%, collapses if it is at most 20\%, and is undecided in between.
It recovers if it survives after falling below 20\%.
Our claims about outcomes rest on runs of 100 steps or more, apart from four 30-step code runs (\cref{tab:code}).

\paragraph{Terms.}
\Cref{tab:roles} gives the roles in each task.
An honest verdict is UNDERIVABLE on an underivable instance or BUGGY on a buggy program.
An attempt is an honest verdict together with its certificate, and it is valid if the checker accepts the certificate.
We write $p_t$ for the rate of valid attempts during training, measured in the same way as $p$, so that $p$ is its value before RL.
When this rate is read off the training samples instead, we call it the validity of attempts in training.

\section{The direction of a step}
\label{sec:step}

The bluff's verdict is also the right answer on the other class (CORRECT on correct programs, DERIVABLE on derivable instances).
So updates on the other class push this shared verdict in the same direction as the bluff does.
To make this precise, we work to first order in the step size and follow the log-odds of the verdict.

On item $i$, let $h_i$ be the probability of the answer UNDERIVABLE or BUGGY, and let $x_i$ be its log-odds.
The fine-tuned bases begin each answer with the verdict, so the part of every answer's gradient that comes from the verdict points along $\nabla_w x_i$.
Summed over the group of samples on an item, the part of the update that comes from the verdict is therefore $P_i\,\nabla_w x_i$, which we call the group's push, with
\[
\mathbb{E}P_i=(n-1)\,h_i(1-h_i)\,(\mu_B-\mu_C),
\]
where $\mu_B-\mu_C$ is the gap between the expected rewards of the two verdicts on item $i$ (\cref{lem:group}).
In other words, an item pushes toward the verdict that pays more on it in expectation, and only while its group is likely to hold both verdicts.
To first order, a step then changes the log-odds on any item $j$, in the batch or not, by $\eta\sum_iP_i\langle\nabla_w x_i,\nabla_w x_j\rangle$ (\cref{lem:step}).
This is the first-order account of how an update on one example moves another \citep{ren2025learning}.
The other class enters the step through these inner products.
Our one modeling assumption is to replace each inner product by its average over pairs of the same kind (within a class, or across the two classes).
Under this assumption we obtain the following (proof in \cref{app:shared}).

\begin{restatable}[The direction of a step]{proposition}{propstep}
\label{prop:step}
Train with GRPO with mean-centered advantages and groups of $n\ge2$ independent samples per item, and treat the verdict on each item as a binary choice.
Suppose that to first order an update on one item moves the log-odds on another in proportion to the inner product of their gradients, and that each inner product can be replaced by its average for that kind of pair.
Then for every item $j$ of the class at stake outside the batch, the expected change in $x_j$ after a step is positive iff
\[
\frac{\tilde p\,r}{b}>1+\kappa, \qquad \kappa=s\,\frac{W_o\,c}{W_t\,b},
\]
where:
\begin{itemize}\setlength{\itemsep}{0pt}\setlength{\parsep}{0pt}\setlength{\topsep}{2pt}
\item $h_i$ is the probability of the answer UNDERIVABLE or BUGGY on item $i$, and $x_i$ its log-odds;
\item on the class at stake ($t$, for target) that answer pays $r$ with probability $p_i$ and the bluff pays $b>0$; on the other class ($o$) that answer is wrong and pays $c>0$ less than the right one on average;
\item $W_t$ and $W_o$ are the sums of $h_i(1-h_i)$ over the classes, and $\tilde p$ is the average of $p_i$ weighted by $h_i(1-h_i)$;
\item $s$ is the average gradient inner product across the two classes divided by the average within the class at stake, and we assume both averages are positive, as they are in every model we measured.
\end{itemize}
With $s=0$ or $W_o=0$ the condition is $\tilde p\,r/b>1$.
\end{restatable}

In words, $\kappa$ is the push from the other class relative to the push on the class at stake, scaled by how far an update on one class moves the verdict on the other.

On code the weighted rate is lower than the plain rate $p$ (0.67 against 0.735 for the coder, and 0.67 against 0.708 for Phi).
With the couplings measured in \cref{app:shared}, $1+\kappa$ is 1.9 to 2.0 in units of $\tilde p$, so in terms of the plain rate it sits near $pr/b=2.1$.
The Datalog bases start almost fully honest, so few items carry weight in $\tilde p$.
On Datalog we therefore use the plain rate for the threshold, and we read $\tilde p$ off the training samples only when we follow a run through training (\cref{sec:dip,app:finite}).

The proposition gives the direction of a step at the current $h$ and $\tilde p$.
Where a run ends depends on how two of its terms move during training.
A checker raises $\tilde p$ (\cref{sec:race}), and $\kappa$ falls as the other class settles (\cref{sec:dip}).

\section{The baseline: when bluffing beats trying}
\label{sec:threshold}

\paragraph{The sign of the push.}
We start with the case $\kappa=0$, in which \cref{prop:step} reduces to a comparison that can be read off a single group.
On an underivable instance, answering DERIVABLE is the bluff.
No valid proof exists, so the bluff pays $b$ for certain, while an attempt at a refutation pays $r$ with probability $p$ and 0 otherwise.
Consider a GRPO group of $n$ samples on such an instance, of which $k$ bluff and $n-k$ attempt.
With advantages centered on the group mean, the expected advantages are
\begin{equation}
A_{\mathrm{bluff}} = \frac{n-k}{n}\,(b - pr), \qquad A_{\mathrm{attempt}} = \frac{k}{n}\,(pr - b),
\label{eq:adv}
\end{equation}
so in every group that holds both a bluff and an attempt, the push toward the bluff has the sign of $b-pr$.
The baseline we test is that this comparison alone decides where a run ends:
\begin{equation}
\boxed{\;\text{honesty survives} \iff \frac{pr}{b} > 1\;}
\label{eq:rule}
\end{equation}
We have assumed advantages centered on the group mean.
Dividing them by the group's standard deviation, as standard GRPO does, moves the threshold by an amount that depends on the group size and on how the per-instance success rates spread around $p$.
On our depth-2 base it would move the threshold down to $r=b$ (\cref{app:std}).

\paragraph{Seven conditions.}
We test the baseline in seven conditions.
On Datalog we compare bases with different $p$, at proof depths 2 and 3, and we also move $p$ within one model by sharpening it, i.e.\ training it first under a reward that pays only valid certificates.
On code we use two models.
For abstention we use a standard reward that pays 1 for a correct answer, 0 for ``I don't know'', and $-\lambda$ for a wrong answer.
Adding $\lambda$ to every payoff puts it in our form, with the sure payoff $b=\lambda$ for abstaining and $r=1+\lambda$ for a right answer.
So the threshold $pr/b=1$ is $\lambda^*=p/(1-p)$, which is Chow's rule.
Here the sure thing is ``I don't know'', as in Chow's own setting, and collapse means claiming ignorance falsely.
In each condition we place runs on both sides of $pr/b=1$, and we bracket the threshold between the nearest ratio at which every checked run collapses and the nearest at which every checked run survives (\cref{tab:cliff}).
\Cref{app:threshold} gives the runs condition by condition.

\begin{table}[t]
\caption{Where the threshold falls.
For each condition, the base's rate $p$ and the bracket in $pr/b$, from the nearest ratio at which every checked run collapses to the nearest at which every checked run survives, with the number of seeds at each end.
These runs are 100 steps long, except those at 0.95 on the depth-2 base and at 1.07 on the sharpened depth-3 base, which ran for 160.
No run of the sharpened depth-3 base has collapsed (--).
Runs inside the brackets, where seeds split or a run is undecided, are in \cref{app:runs}.}
\label{tab:cliff}
\centering\scriptsize
\setlength{\tabcolsep}{3pt}
\begin{tabular}{@{}llcccc@{}}
\toprule
 & & & \multicolumn{2}{c}{$pr/b$ bracket} & \\
\cmidrule(lr){4-5}
task & base & $p$ & collapses & survives & seeds \\
\midrule
Datalog & depth 2 & 0.77 & 0.80 & 0.95 & 2, 4 \\
 & depth 2, sharpened & 0.86 & 0.60 & 1.10 & 1, 1 \\
 & depth 3 & 0.66 & 0.80 & 1.07 & 1, 2 \\
 & depth 3, sharpened & 0.77 & -- & 1.07 & --, 3 \\
code & 3B coder & 0.735 & 0.95 & 2.00 & 1, 1 \\
 & Phi-3.5 & 0.708 & 0.95 & 2.00 & 1, 1 \\
abstention & GSM8K & 0.701 & 0.78 & 0.94 & 2, 1 \\
\bottomrule
\end{tabular}
\end{table}

\paragraph{The measured thresholds.}
With no constant fitted, no run above $pr/b=1$ ends collapsed in any condition, and none of the six brackets with both ends lies above 1.
Four contain 1, and two lie below it, namely the brackets of the depth-2 base (0.80--0.95) and of abstention (0.78--0.94).
Of the four, the bracket of the sharpened depth-2 base (0.60--1.10) is too wide to say much.
The bracket of the sharpened depth-3 base has only an upper end, at 1.07, because its one run below 1 is undecided.
Sharpening also moves the threshold within one model.
At the same payoff ratio, $r/b=1.39$, the depth-3 base ($pr/b=0.92$) survives in none of three seeds, while the sharpened model ($pr/b=1.07$) survives in all three (\cref{fig:cut}(b)).
The two start about equally honest (94.2\% and 96.5\%), but the sharpened model has also been trained under the checker already, so its other class has settled as well (\cref{sec:dip}).
The collapses at the lower edges of the brackets look permanent, with at most 3\% honest verdicts left at their last evaluation.
The baseline fails in two ways, both of which \cref{prop:step} allows once $\tilde p$ and $\kappa$ move during training.
The two brackets below 1 are where checked runs won the race.
The depth-2 base survives at 0.93 (one of two seeds) and at 0.95 (four of four), and the abstention run at 0.94 keeps answering (\cref{sec:race,app:abstention}).
We ran no lottery for abstention.
The code brackets are wide because the runs at 1.25, between 1 and $1+\kappa$ (about 2.1 on code), first fall nearly to zero and are still climbing back when training ends (\cref{sec:dip}).
The rest of the paper follows these two effects, and a third that only the coupling predicts, in which a reward that pays the honest verdict too much hurts the other class (\cref{sec:toomuch}).

\section{Training moves the threshold through $p_t$}
\label{sec:race}

The first term of \cref{prop:step} that moves during training is the success rate.
A checker pays only valid attempts, so training under it raises the rate.
The lottery, our control, pays at the base's success rate however the attempts change.

\paragraph{The lottery control.}
We used the checker to grade every attempt in eleven matched pairs of checked and lottery runs, including the attempts of the lottery runs, although their reward ignores validity.
Under the checker the validity of attempts in training rises in every run, while under the lottery it stays near the base's rate for as long as attempts are common.
Where the lottery runs' attempts stay common to the end, on the depth-2 base at $pr/b=1.07$ and 3.0, their validity ends at 0.77, while the checked runs reach 0.86 and 0.87 (\cref{app:race}).
Below the threshold the lottery runs' attempts thin out within 40 to 80 steps, so the comparison there rests on few, selected attempts and is noisy.
The rise in the checked runs comes from improvement, and only a little from the model dropping the instances it fails.
At $pr/b=0.95$, the checked runs' attempts at the end of training are valid 12 points more often than the base's attempts, and of this gap selection accounts for about 1 point and improvement for about 11 (\cref{app:race}).
The outcomes of the runs point the same way as the validity.
In the nine matched pairs below the threshold, five checked runs survive, two are undecided, and all nine lottery runs collapse (one-sided Fisher exact test on survival, 0.015).

\paragraph{Recovery below the threshold.}
Below the threshold, every group that holds both a bluff and an attempt favors the bluff, so attempts become rarer.
Under a checker, though, $p_t$ rises.
If it passes $b/r$ while attempts are still common, honesty recovers.
If the attempts die out first, the collapse is permanent.
Just below the threshold, at $pr/b=0.95$, all four checked seeds come back, to 68--85\% honest after 160 steps.
At 0.93, one of two checked seeds survives (\cref{fig:race}(a, b)).
In the checked runs at 0.95, the rate $p_t$ measured on fresh instances rises from 0.77 to 0.86--0.87 by step 160, so $p_t\,r/b$ reaches 1.06--1.07.
Just above the threshold, at $pr/b=1.07$, the lottery run and both checked runs survive (\cref{tab:long}).
On the depth-3 base at $pr/b=0.92$, the three checked seeds fall to 4--15\% by step 40 and end at 20--47\% after 160 steps, while every lottery seed ends at 0--0.2\% (\cref{fig:race}(c)).
The validity of the checked seeds' attempts in training passes $b/r$ between steps 20 and 40, while the seeds fall from 37--51\% to 4--15\% honest, and then they climb back slowly.

\paragraph{Predicting the race.}
Seeds that end differently look alike in the first steps.
The two checked seeds at $pr/b=0.93$ match in honesty and in the validity of their attempts through step 30, and the one that later wins first falls further.
We can still predict the chance of winning.
We built a model of the race from \cref{prop:step} and one further assumption, that the checker raises the validity of attempts at a rate proportional to the share of answers that are attempts, up to a saturation level (\cref{app:racemodel}).
Its three constants are fit only on runs outside the race band, the ratios just below the threshold at which checked runs can win, and its only noise is that of sampling a batch.
For a checked run on the depth-2 base, it predicts a chance of winning of 0 at $pr/b=0.80$, 0.15--0.59 at 0.93 after 100 steps, and 1 at 0.95 after 160 steps (the lengths of those runs).
In the runs at these ratios, 0 of 3 (two seeds, one of them run twice), 1 of 2, and 4 of 4 won.
It also predicts that every lottery run below the threshold collapses.
The range at 0.93 is over two versions of the model's update, which match the fitting runs equally well.
So on this base the model places the race in a narrow band, at about 0.92--0.95, and within that band the outcome of a single run is a matter of chance.
For the depth-3 base at $pr/b=0.92$, we fit the same model on depth-3 runs at other ratios and recorded its prediction before the runs at 0.92 were done.
It gave each checked seed a chance of 0.04--0.13 of winning within 160 steps, and it predicted that every lottery seed would collapse.
No checked seed won, and every lottery seed collapsed (\cref{fig:race}(c)).
The model also predicted that the checked seeds, though they would not win, would stay above the lottery seeds, and from step 60 on every checked seed is more honest than every lottery seed.
\Cref{app:racemodel} compares the runs with the predicted trajectories.

\section{The early dip}
\label{sec:dip}

\begin{figure*}[t]
\centering
\includegraphics[width=\textwidth]{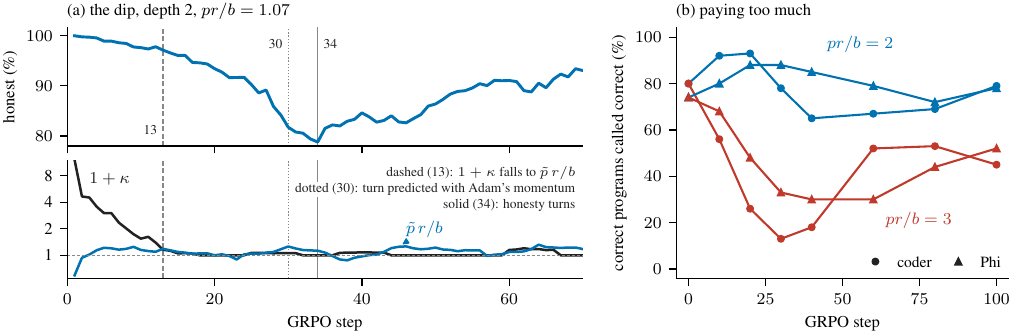}
\caption{The other class, which shares the bluff's verdict, sets how a run starts and where the payoff ratio stops helping.
(a) The depth-2 base at $pr/b=1.07$ (seed 1).
Top: honest verdicts on underivable instances in the training samples (7-step mean).
Bottom: the two sides of \cref{prop:step}, $1+\kappa$ and $\tilde p\,r/b$, read off the training samples in 7-step windows, with the coupling $s=1.03$ measured on the base.
The dashed line marks step 13, where $1+\kappa$ first falls to $\tilde p\,r/b$.
The dotted line marks the turn predicted by summing the push of each step with Adam's momentum (step 30), and the solid line the turn of honesty (step 34).
(b) The share of held-out correct programs called correct by both code models at $pr/b=2$ and 3.
The ceiling predicted from the couplings lies between them, at about 2.15--2.2 (\cref{sec:toomuch}).}
\label{fig:dip}
\end{figure*}

The second term of \cref{prop:step} that moves during training is $\kappa$.
As the other class settles, $W_o$ shrinks and $\kappa$ falls.
By \cref{prop:settle}, if $ss'<1$ and $1<\tilde p\,r/b<1+\kappa$, each step moves the other class toward its right answer.
So the comparison of $\tilde p\,r/b$ with $1+\kappa$ sets the direction in which a run starts, and the comparison of $p_t\,r/b$ with 1 sets where it ends.
This gives three regimes:
\begin{itemize}\setlength{\itemsep}{0pt}\setlength{\parsep}{0pt}\setlength{\topsep}{2pt}
\item below 1, a run collapses unless training raises $p_t$ in time (\cref{sec:race});
\item between 1 and $1+\kappa$, it first falls and then rises, if its attempts last;
\item above $1+\kappa$, it rises from the start.
\end{itemize}
\paragraph{The dip on Datalog.}
On the depth-2 base the coupling, measured from the gradients of the verdict's log-odds as on code, is $s=1.03$ (\cref{app:shared}).
A coupling near 1 means that before RL an update moves the verdict on every item almost alike, as if it shifted one bias shared by both classes, so the push from the other class reaches the class at stake almost undiminished.
In the training samples of the five checked runs from this base at $pr/b=1.07$, 1.25, and 1.6, $1+\kappa$ starts above $\tilde p\,r/b$ and falls to it between steps 11 and 18, as the derivable instances settle (\cref{fig:dip}(a) and \cref{app:finite}).
Until then the push favors the bluff.
On derivable instances the model is quickly taught to stop answering UNDERIVABLE, and this pull toward DERIVABLE carries over to underivable instances, where bluffs rise.
Every checked run at $pr/b=1.07$ and 1.25 falls from 99\% honest before RL to 84--94\% at its first evaluations, and then climbs back to 94--98\%.
The checked runs at 1.6--3.0 stay at 97--100\% at every evaluation (\cref{tab:long}), though their training samples also dip a few points (\cref{app:finite}).
This is as expected, since on the base $1+\kappa$ is about 3.7, above every ratio we ran (\cref{app:shared}).
On Datalog the gap $c$ is about 1 whatever $b$ is, so $\kappa$ grows as $b$ shrinks.
The condition $\tilde p\,r/b>1+\kappa$ is then in effect a condition on the difference $\tilde p\,r-b$.

\paragraph{The timing.}
\Cref{prop:step} gives the direction of one step, and honesty turns only once enough steps point the same way.
In the training samples honesty keeps falling after $1+\kappa$ reaches $\tilde p\,r/b$, and it turns at steps 29--34.
Two things delay the turn.
First, once the margin $\tilde p\,r/b-1-\kappa$ turns positive it stays near 0 for a while, so those steps push little either way (\cref{fig:dip}(a)).
Second, our optimizer, Adam, moves the weights along a running average of past gradients (its momentum), and this carries the early pull toward bluffing forward.
Summed through this momentum, the push of \cref{prop:step} predicts the turn 4 to 8 steps before honesty turns.
This gap is positive in all five runs, but it is within the sampling noise in four of them and at its edge in the fifth (\cref{app:finite}).
So the step condition, with momentum, places the turn of a run to within a few steps.

\paragraph{Removing the other class.}
On code $c=b$, so $\kappa$ does not depend on $b$.
With the couplings measured on the fine-tuned models, $1+\kappa$ is about 2.1 in units of $p$ for both models.
This is why runs just above $pr/b=1$ first fall, while runs above $1+\kappa$ rise from the start (\cref{app:shared}).
Training the coder at $pr/b=1.25$ on buggy programs alone sets $\kappa=0$, since the other class is then absent, and in this single run the dip goes away (\cref{tab:code}).
BUGGY recall is 92\% at step 10, against 26\% when correct programs are in the RL set.
However, with nothing to hold the correct programs in place, the coder ends up calling them buggy too.
By step 30 it calls 0\% of them correct, against 80\% before RL.

\paragraph{The dip and the race.}
The dip also explains why the checker lowers the threshold so little.
Below the threshold, the dip adds to the bluff's advantage in the first steps, so attempts thin out before the checker has had time to raise $p_t$, and on the depth-2 base the threshold moves from 1 only to between $pr/b=0.80$ and 0.95.
A model whose other class has already settled has no dip, so it keeps attempting and gives the checker more time.
Training first at a generous ratio, where the run survives, settles the other class and, under the checker, raises $p_t$, so a model cut to a lower ratio afterward starts without the dip and closer to its threshold.
We test this schedule in \cref{sec:cut}.

\begin{figure*}[t]
\centering
\includegraphics[width=0.83\textwidth]{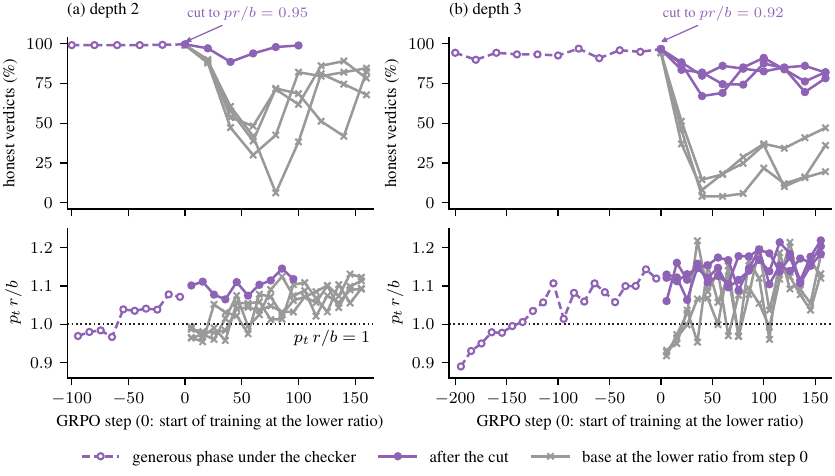}
\caption{A generous phase moves the threshold.
For every run, step 0 is the start of training at the lower ratio, and the generous phase is drawn dashed at negative steps.
(a) In purple, the depth-2 base is trained for 100 steps at $pr/b=3.0$ and then cut to $pr/b=0.95$ for 100 more steps, under the checker throughout.
In gray, the same base is trained at $pr/b=0.95$ from the start (four seeds, 160 steps).
(b) In purple, the depth-3 base is sharpened, i.e.\ trained for 200 steps under a reward that pays only valid certificates, and then trained at $pr/b=0.92$ for 160 steps (three seeds from the one sharpened model).
In gray, the depth-3 base is trained at $pr/b=0.92$ from the start (three seeds, 160 steps).
All ratios use the base's $p$.
The top panels show honest verdicts at each evaluation.
The bottom panels show $p_t\,r/b$ at the lower ratio throughout, including the generous phase.
Here $p_t$ is read from the validity of attempts in training (graded by the checker, in 10-step bins with at least 20 attempts), while the text measures it on fresh instances.
The dotted line is the threshold $p_t\,r/b=1$.}
\label{fig:cut}
\end{figure*}

\section{Paying too much}
\label{sec:toomuch}

The baseline says that raising $r/b$ can only help honesty, but through the coupling it can also hurt the other class (\cref{app:shared}).
The push that a step gives the class at stake also reaches the other class, through the coupling in the other direction, which we write $s'$.
On code, $s'$ measures how far an update driven by buggy programs moves the verdict on correct ones.
There the verdict BUGGY is wrong, and the correct programs' own groups pull it back toward CORRECT with strength $W_oc$.
The push on the other class is therefore $s'W_t(\tilde p\,r-b)-W_oc$, and above the ceiling
\[
\frac{\tilde p\,r}{b}=1+\frac{W_o\,c}{s'\,W_t\,b}
\]
the push from the buggy programs wins, and the model starts calling correct programs buggy.

On code the couplings measured on the starting models put the ceiling at about $pr/b=2.15$--2.2 for both models, just above $1+\kappa$ (\cref{app:shared}).
The two are close because $ss'$ is near 1 on code (0.95 for the coder and 0.93 for Phi).
So on code there is almost no ratio at which honesty rises from the start without the model also starting to call correct programs buggy.
We trained each model once at $pr/b=2$, below both points, and once at 3, above both, so these runs test the two points together (\cref{fig:dip}(b)).
At 2, both models call correct programs correct about as often at step 100 as before RL (79\% and 78\%, against 80\% and 74\%), although the coder dips to 65\% at step 40.
At 3, that share falls to 13\% for the coder and 30\% for Phi within 40 steps, and it is still only 45\% and 52\% at step 100 (\cref{fig:dip}(b) and \cref{tab:codecheck}).
Meanwhile BUGGY recall at 3 rises from the start, to 93\% and 87\% at step 100.
So the best ratio overall lies below the one that makes the honest verdict most attractive on buggy programs.

On Datalog we see at most a small effect.
On the depth-2 base the share of held-out derivable instances answered DERIVABLE ends only a point or two lower at the highest ratios than at $pr/b=1.07$ (\cref{sec:setup,app:runs}).
There the right answer on the other class pays $c\approx1$, more than $b$, which strengthens that class's own pull.
On the depth-2 base the measured couplings put the ceiling near $pr/b=4$, above every ratio we ran, and after the first 13 to 18 steps the training samples hold too few derivable instances that still draw both verdicts to follow it further (\cref{app:shared,app:finite}).

\section{Generous, then cut}
\label{sec:cut}

The race suggests a schedule.
Consider a model that has already improved, and whose other class has already settled, so that $\kappa$ is near 0.
\Cref{prop:step} predicts that such a model can afford a lower ratio, since the comparison applies at every step with the current model's success rate $p_t$ in place of the base's $p$.
This gives the recipe boxed on page~\pageref{box:recipe}.
We tested this on both Datalog bases (\cref{fig:cut}).
At depth 2 we trained the base for 100 steps at $pr/b=3.0$ under the checker and then cut the ratio to $pr/b=0.95$, with $p$ the base's rate.
At the cut, the trained model's rate measured on fresh instances is $p_t=0.866$, so the new ratio gives it $p_t\,r/b=1.06$.
The trained model stays honest for 100 more steps, at 89--99\% in every evaluation.
Bluffs come back after the cut and lose to attempts (\cref{app:cut}).
The base trained at that ratio from the start falls to 6--48\% honest in all four seeds before it comes back to 68--85\% by step 160.
So in this run the schedule avoids the dip of the runs trained at that ratio from the start.
When the generous phase is run under the lottery instead, $p_t$ stays at 0.76, and after the same cut, at $p_t\,r/b=0.93$, the model dips to 29\% before it ends 83\% honest (\cref{app:cut}).
The check during the generous phase thus avoids the dip.
At depth 2 the window of ratios with $pr/b<1$ but $p_t\,r/b>1$ is narrow, because the generous phase raises $p_t$ only from 0.77 to about 0.87.

At depth 3 sharpening raises $p$ from 0.66 to 0.77, which leaves a wider window.
Sharpening is itself a generous phase under the checker, since it pays only valid certificates ($b=0$).
We used it as the generous phase and cut the sharpened model to $r/b=1.39$, where the base sits at $pr/b=0.92$ and the sharpened model at 1.07, and compared it with the base trained at that ratio from the start (\cref{fig:cut}(b)).
In three seeds each, the sharpened model stays at least 67\% honest for 160 steps and ends at 78--82\%.
The base falls to 4--15\% and ends at 20--47\%, with one seed collapsed and two undecided.
So at depth 3 the schedule avoids more than a dip.

\section{The collapse is bluffing}
\label{sec:frozen}

\begin{figure}[t]
\centering
\includegraphics[width=\columnwidth]{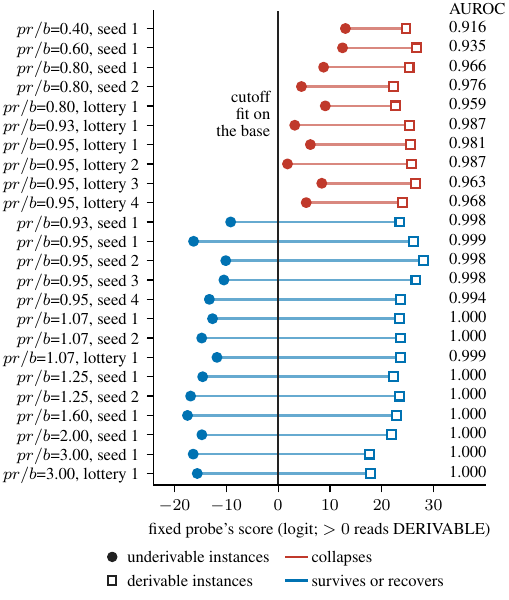}
\caption{The probe fit on the base, held fixed.
The plot shows the probe's mean score on held-out underivable (circles) and derivable (squares) instances at the base's chosen layer (27), for the depth-2 runs that collapse (red) and survive (blue), with the probe's AUROC.
Scores above 0 read DERIVABLE under the cutoff fit on the base.
In the collapsed runs the underivable scores have moved up past 0, while the two classes stay apart.}
\label{fig:frozen}
\end{figure}

On depth-2 Datalog, when honesty collapses, the model stops reporting the truth but still represents it.
We show this with a linear probe, i.e.\ a linear classifier on the model's activations.
We fit it on the base model's activations at a chosen layer before RL, and then hold it fixed (\cref{app:frozen}).
We apply it to the ten collapsed models trained from the depth-2 base.
The probe still separates the classes, i.e.\ underivable instances still score below derivable ones.
Its AUROC (the chance that a random derivable instance scores above a random underivable one) is 0.92--0.99 at the base's chosen layer.
The underivable scores move up, past the cutoff for most instances.
So with its original cutoff the probe calls 55--98\% of underivable instances derivable, as the model now answers (\cref{fig:frozen}).
In the runs that end honest, including those that fell and recovered, the probe still classifies correctly with its original cutoff (91--99.6\% accuracy).
The collapse thus moves the underivable instances along the probe's direction and leaves the distinction between the two classes intact.
The distinction survives the collapse, and it is still in place in the runs that fell and then returned to honesty.

\section{Related work}
\label{sec:related}

We see five things in this paper as new: (1) the direction of a GRPO step, in which the shared verdict enters through a coupling that we measure from gradients, which explains the early dip and predicts a ceiling on the payoff ratio; (2) the threshold stated in terms of certificates, with $p$ measured before RL and tested in seven conditions; (3) the race, in which a checker raises $p_t$ and so lowers the threshold, while a lottery with the same mean does not; (4) starting generous and then cutting the ratio, with a criterion for when to cut; and (5) a probe direction fit before RL that still reads the truth after a collapse.

Our analysis of a step takes the first-order account of how an update on one example moves another \citep{ren2025learning} and applies it to GRPO's group advantages and to a verdict that the two classes share.
Several pieces of the race are already known (\cref{app:related}).
GRPO drains probability from correct actions that it does not sample.
The honest action is an arm of a rested rising bandit, i.e.\ one whose payoff rises when it is pulled.
Over-abstention can be transient or permanent.
Finally, others have built probes that read the truth from a model's activations when its output does not, and we discuss this work at the end of this section.

\paragraph{Reward thresholds and collapse laws.}
\citet{kalai2025why} state for language models that binary grading rewards guessing, and that a penalty for wrong answers sets a confidence threshold.
This is Chow's reject rule \citep{chow1970optimum,cortes2016rejection}.
\citet{wei2025truthrl,jha2026humility,dou2025hesitation,wu2025behavioral} sweep such penalties or measure refusal under RL, and \citet{damani2025beyond} add a Brier score on stated confidence to the correctness reward.
In these works the threshold, where there is one, is on a confidence that cannot be observed.
Ours is on a success rate that the checker measures, because the honest action can be verified.
\citet{song2025hallucination} find that RL fine-tuning lowers refusal on unanswerable math problems, and that adding such problems to training, with refusal rewarded only on them, restores it.
In concurrent work, \citet{arora2026graceful} observe that RL can learn a useful policy when the reward for a safe fallback is below the policy's success rate.
They trace the collapse onto the fallback to vanishing gradients near saturation.
In their setting the sure option is a safe fallback that they want the model to learn, and the boundary is found by sweeping its reward.
In ours the sure option is the bluff, and the boundary is predicted from $p$ measured before RL.
\citet{li2026audit} likewise use the sign of the mean-centered GRPO push at the starting checkpoint to predict whether an audit design suppresses reward hacking, and the prediction is right in 33 of 39 runs.
Also in concurrent work, \citet{li2026multireward} prove the equal-payoff case of the comparison of $pr$ with $b$, in which GRPO converges to whichever of two equally weighted reward channels is obtained more reliably.
\citet{pan2026tiar} derive an abstention reward from the model's sampled accuracy.
They note that, placed in the reward, it would enter the group mean and standard deviation, so they add it to the advantage after normalization.
\Cref{app:std} discusses how standardized advantages move the threshold.
\citet{rad2026rate} model GRPO as a replicator process in which a noisy verifier reverses learning if and only if its false-positive rate exceeds its true-positive rate.
Our threshold is their sign law with the payoff for the unverified claim in the place of the false positive.
In our setting the reward pays this false positive by design \citep[cf.][]{cai2025noisy,zhang2026leaky}.
Their law also explains why the lottery cannot raise $p_t$, since it pays valid and invalid attempts alike.
Their experiments hold the verifier's error rates fixed, and they leave rates that drift during training to future work.
Here the honest action's success rate drifts in this way, rising as the policy improves.
\citet{dou2025hesitation} see over-abstention that is transient under one penalty, and a collapse to universal abstention under a harsher one.
They read the transient as the model learning where to abstain.
The race offers a mechanism for that pattern, and the lottery gives a way to test it causally.
Language agents lock into exploits before they find safer strategies \citep{cagatan2026gridworlds}, and in our setting the race quantifies that competition.

\paragraph{Lenient-then-strict schedules.}
Starting generous and tightening later is an old idea.
In shaping, a trainer reinforces successive approximations to the target behavior \citep{peterson2004shaping}.
Classifiers that can abstain are first trained without the abstain option \citep{ziyin2019gamblers,thulasidasan2019abstention}, and robot learning moves from a dense shaped reward to the sparse true one \citep{luo2020dense2sparse}.
In RL for language models, partial test credit speeds early GRPO learning in code generation without reliably raising the final pass rate \citep{li2026passrate}.
KARL separates an exploration stage from the stage that calibrates abstention, after finding that a static ternary reward locks in early over-abstention \citep{gao2026karl}.
AWA-RL adapts the abstention reward to training dynamics \citep{zhang2026awa}.
Curricula over task difficulty change the problems and keep the reward.
They move from easy problems to hard ones \citep{parashar2025e2h}, or are gated on competence and paired with an abstention reward \citep{li2025rlaar}.
These schedules decide by heuristic when to tighten and by how much.
The exception is the online reward-tuning baseline in the concurrent work of \citet{arora2026graceful}, which keeps the fallback reward at a fixed multiple of the running success rate.
The threshold gives a criterion for when to cut.
\Cref{prop:step} predicts that a target ratio is safe once the trained model has $p_t r/b>1$, and this can hold where the starting model has $pr/b<1$.

\paragraph{Internal state vs output.}
Earlier work shows that a model's internal activations decode the truth when its output is wrong, both for instructed lying \citep{campbell2023localizing} and for false or hallucinated statements \citep{azaria2023internal,orgad2024know}.
\citet{shi2025know} show that supervised fine-tuning (SFT) damages the model's expression of ``I don't know'', while probes still separate answerable from unanswerable questions, including probes fit on the base model and applied unchanged to the fine-tuned one.
Our frozen probe gives the same kind of evidence after RL, in models whose output has collapsed to one verdict.
When a model is prompted or fine-tuned to sandbag, i.e.\ to hold back a capability on purpose \citep{vanderweij2024sandbagging}, the withheld capability can come back when its activations are edited \citep{tan2026causal,le2026grafting}.
After RLVR, a probe for the task's own domain, refit on the trained model, still separates hacked code from honest code, while refit probes for deception in general do much worse \citep{taufeeque2026obfuscation} (see \citealp{mallen2023quirky} for probes after SFT).
\citet{nabi2026rotate} finds that GRPO keeps the accuracy of a refit probe while rotating its direction, so a probe held fixed from before RL gives separate evidence.
\citet{fel2026shaping} find that a probe fit before training stays largely reliable afterward even when its score is used in training, as long as no gradient passes through it, and we use ours only to read the trained models.

\Cref{app:related} covers further work on training dynamics (the absorbing branch of the race, rising bandits, random rewards), on what RL changes in the representation, and on reward hacking, abstention, and refutation.

\section{Limitations}
\label{sec:limits}

We trained models with 1.5B to 3.8B parameters for 30 to 200 GRPO steps, on one synthetic task, one code task, and grade-school math.
Our analysis of a step is first order, replaces the gradient inner products by class averages, and gives only the direction of a step.
The timing of a run also depends on the optimizer, and even with Adam's momentum included, the predicted turn of honesty comes 4 to 8 steps early (\cref{sec:dip}).
The Datalog results come from one 3B model on two tasks, whose proofs are two and three levels deep.
For both code models we bracket the threshold only loosely, between $pr/b=0.95$ and 2.0, because their runs at 1.25 are still climbing when training ends.
Most bracket ends rest on a single seed, and every code run has one seed (\cref{tab:cliff,tab:code}).
Near the threshold, runs that differ only in their seed reach different outcomes, and our model predicts only the odds that a run wins (\cref{sec:race}).
The assumption in our model of the race about how the checker raises the validity of attempts is fit to data.
Below the threshold the lottery runs' attempts thin out early, so there the comparison of validity under the checker and under the lottery rests on few, selected attempts.
We tested the generous-then-cut schedule from a single generous model at each depth, and at depth 3 the generous phase was sharpening.
The thresholds we measure are for bases trained at one ratio from the start.
We expect that later in training, when attempts last longer, a run under the checker recovers from further below the threshold.
Our probe is linear and is read at one token position.

\section{Conclusion}

To first order, a GRPO step moves honesty up or down according to the sign of $\tilde p\,r/b-1-\kappa$.
Here the pull of the other class, $\kappa$, sets how a run starts, and it fades as that class settles.
Once it has faded, one comparison remains, between the payoff for a verified certificate times the model's success rate at producing one, and the payoff for the unchecked verdict.
If the success rate stays fixed, this comparison is the threshold.
The coupling between the two classes that sets $\kappa$ also sets a ceiling, above which the push toward the honest verdict spills over onto the other class.
A checker lets the attempts improve, so a run that starts slightly below the threshold can still win.
Under the lottery, which has the same mean payoff but no check, no run that started below the threshold won.
A model of the race predicts the odds of winning, and the outcomes of the depth-3 runs came out as it predicted before they were done.
A model first trained generously under the checker stays honest at a harsher reward where a model trained at that reward from the start falls, in one depth-2 run and in three depth-3 seeds.
For reward design, we suggest three things.
First, keep $pr/b$ above 1 with a margin but below the ceiling, and measure the success rate on the model being trained.
Second, pay generously, with the check, before cutting.
Third, keep the check, since it lets the model improve.

\section*{Impact Statement}
This paper studies when reinforcement learning from verifiable rewards teaches a model to make claims it cannot support.
It gives a way to check, before training, whether a reward will do so, and a schedule that avoids it.
We expect these to help practitioners design rewards that keep models honest.
The same analysis says which rewards would train bluffing, but such rewards are easy to find without it.
Our experiments use small open models on a synthetic task, a code task, and grade-school math.
Whether the threshold holds at larger scale and on open-ended tasks is untested.

\section*{Acknowledgments}
We thank Indranil Halder and Sanjeev Arora for useful discussions.

This work was supported by the Defense Advanced Research Projects Agency (DARPA) through the Super Intelligence Quantified (SIQ) program, under Cooperative Agreement No.\ HR00112520028.
The views, opinions, and/or findings expressed are those of the authors and should not be interpreted as representing the official views or policies of the Department of Defense or the U.S.\ Government.

\section*{Use of AI assistance}
An AI assistant (Claude, Anthropic) was used to assist with experiments and analyses, and to help with the writing.
The authors take full responsibility for the content.

\bibliography{refs}
\bibliographystyle{icml2026}

\newpage
\appendix
\onecolumn
\counterwithin{proposition}{section}
\counterwithin{lemma}{section}
\renewcommand{\topfraction}{0.9}\renewcommand{\bottomfraction}{0.8}\renewcommand{\textfraction}{0.07}\renewcommand{\floatpagefraction}{0.8}
\setcounter{topnumber}{4}\setcounter{bottomnumber}{2}\setcounter{totalnumber}{6}
\makeatletter\setlength{\@fptop}{0pt}\setlength{\@fpbot}{0pt plus 1fil}\makeatother
\FloatBarrier
\section{The threshold in each condition}
\label{app:threshold}

\begin{figure*}[htbp]
\centering
\includegraphics[width=\textwidth]{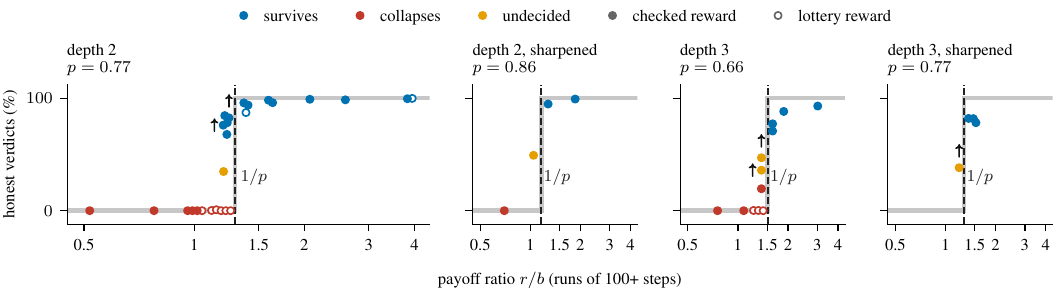}
\caption{Where honesty breaks is set by $p$, measured before RL.
Each panel shows honesty at the last evaluation of every run of 100 steps or more on one Datalog base, against the payoff ratio $r/b$.
We plot $r/b$ so that each base's predicted threshold $1/p$ sits at its own place.
Colors give the outcome as in \cref{fig:regimes}, with filled markers for checked runs and open ones for lottery runs.
Each base's honesty turns near its own $1/p$ (dashed line, with the predicted step drawn faint).
Sharpening raises $p$ from 0.77 to 0.86 at depth 2 and from 0.66 to 0.77 at depth 3, so the panels also move $p$ within one model.
Up-arrows mark runs that end below 90\% but at least 10 points above their value at the evaluation 40 steps earlier.
Points just left of their base's dashed line that end honest are runs that fell and then recovered by improving their success rate (\cref{sec:race}).}
\label{fig:law}
\end{figure*}

\begin{figure}[htbp]
\centering
\includegraphics[width=0.5\textwidth]{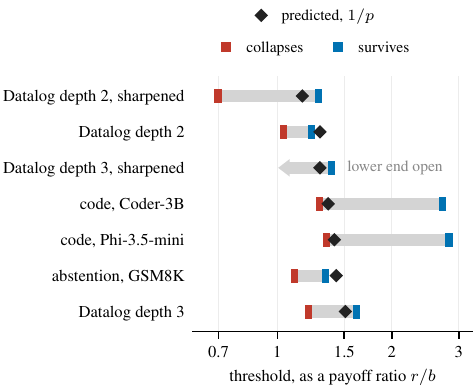}
\caption{Predicted and measured threshold for every condition, as a payoff ratio $r/b$, ordered by the prediction.
The diamond is the prediction $1/p$, with $p$ measured before RL, which is the threshold if the model's success rate stays fixed.
The gray bracket is the measured threshold, from the nearest ratio at which every checked run collapses to the nearest at which every checked run survives (\cref{tab:cliff}).
An arrow marks an end that no run has reached.}
\label{fig:predobs}
\end{figure}

\paragraph{Datalog.}
We train the depth-2 base with RL on the task whose proofs are two levels deep, and we place its runs on either side of its own threshold.
It collapses at $pr/b=0.40$ and 0.60, and at 0.80 in seed 1 (run both on our GPUs and on the second cluster) and in seed 2.
It survives at 1.07 and 1.25 (two seeds each), 1.6, 2.0, and 3.0 (\cref{fig:regimes,fig:law}).
Sharpened to $p=0.86$, the same model collapses at 0.60 and survives at 1.10 and 1.60.
Its run at 0.90 is undecided, at 49\% honest and still falling at step 100.
The depth-2 base also survives below the threshold, at 0.93 (one of two seeds) and at 0.95 (four of four).
The race explains these runs (\cref{sec:race}).
The depth-3 base, which we train on the task whose proofs are three levels deep, has $p=0.66$.
It collapses at 0.50 and 0.80, and survives at 1.07 (two seeds), 1.25, and 2.0.
At 1.07 the two seeds dip to 56--57\% and end at 71\% and 77\%.
At 1.25 it is 74\% honest at step 20 and 88\% at step 100, while $p_t$ rises to 0.76.
At 2.0 it is 86\% honest at step 20 and 93\% at step 100.
At 0.92 we ran three checked and three lottery seeds for 160 steps.
Every lottery seed collapses.
One checked seed ends collapsed, at 19.5\%, and two are undecided, at 47\% and 36\% (\cref{fig:race}(c)).
Their $p_t$ on fresh instances ends at 0.76--0.78.
Sharpened to $p=0.77$, the depth-3 base survives at 1.07 in three seeds of 160 steps, ending at 78--82\%.
Its run at 0.93 is undecided, at 38\% honest at step 100, so its bracket has no lower end.

\paragraph{Code.}
Both code models collapse at $pr/b=0.95$ and below, and survive at 2 and 3.
In between, at 1.25, both first fall, the coder to 0\% and Phi to 7\%, and are still climbing back when training ends.
The coder reaches 39\% at step 100 and Phi 49\% at step 200, and Phi at 1.6 reaches 58\% (\cref{tab:code}).
We explain in \cref{sec:dip} why runs just above $pr/b=1$ first fall.

\paragraph{Abstention.}
The abstention runs are in \cref{app:abstention}.

\Cref{fig:predobs} compares the predicted and measured thresholds of every condition.

\FloatBarrier
\section{Every run}
\label{app:runs}

\Cref{tab:long} lists every Datalog run with $b>0$, and \cref{tab:code} lists every code run.
Outcomes follow the definitions of \cref{sec:setup}.
On 400 instances, a recall near 50\% has a standard error of 2.5 points, and a recall within 5 points of 0 or 100\% has a standard error of at most about one point.
On the 337 programs of the code task, the standard error is up to 2.7 points.
On the depth-2 base, the share of held-out derivable instances answered DERIVABLE at the last evaluation is 100\% at $pr/b=1.07$, 99.0\% at 2.0, and 98.0\% at 3.0.

\newcommand{\TabLongCaption}{Every Datalog run with $b>0$, with its underivable recall at each evaluation and its outcome at the last evaluation.
The ratio $pr/b$ uses the base's $p$, measured as in \cref{sec:setup}, which is 0.77 and 0.66 at depths 2 and 3, and 0.86 and 0.77 after sharpening.
The runs after a generous phase start from the depth-2 base trained for 100 steps at $pr/b=3.0$, under the checker ($p_t=0.87$) or under the lottery ($p_t=0.76$).
For these two runs the column gives $p_t\,r/b$, with $p_t$ that rate.
Seed 2 at $pr/b=0.80$ was evaluated every 10 steps.
The row marked $^\ast$ reruns seed 1 at $pr/b=0.80$ on our second cluster (\cref{app:training}), as a check of reproducibility.}
{\scriptsize\setlength{\tabcolsep}{3pt}\setlength{\LTcapwidth}{\textwidth}
\begin{longtable}{lllccll}
\caption{\TabLongCaption}\label{tab:long}\\
\toprule
base & $b,\ r$ & reward & $pr/b$ & seed & honest verdicts (\%) at each eval (step: value) & outcome \\
\midrule
\endfirsthead
\toprule
base & $b,\ r$ & reward & $pr/b$ & seed & honest verdicts (\%) at each eval (step: value) & outcome \\
\midrule
\endhead
\bottomrule
\endfoot
depth 2, $p$=0.77 & 0.6, 0.311 & checked & 0.40 & 1 & 20: 65, 40: 0, 60: 0, 80: 0, 100: 0 & collapses \\
 & 0.6, 0.466 & checked & 0.60 & 1 & 20: 59, 40: 0, 60: 0, 80: 0, 100: 0 & collapses \\
 & 0.6, 0.621 & checked & 0.80 & 1 & 20: 85, 40: 1, 60: 0, 80: 0, 100: 0 & collapses \\
 & 0.6, 0.621 & checked & 0.80 & 1$^\ast$ & 20: 89, 40: 2, 60: 0, 80: 0, 100: 0 & collapses \\
 & 0.6, 0.621 & checked & 0.80 & 2 & 10: 95, 20: 81, 30: 12, 40: 1, 50: 0, 60: 0, 70: 0, 80: 0, 90: 0, 100: 0 & collapses \\
 & 0.6, 0.621 & lottery & 0.80 & 1 & 20: 68, 40: 0, 60: 0, 80: 0, 100: 0 & collapses \\
 & 0.6, 0.722 & checked & 0.93 & 1 & 20: 89, 40: 40, 60: 44, 80: 59, 100: 76 & survives \\
 & 0.6, 0.722 & checked & 0.93 & 2 & 20: 91, 40: 67, 60: 35, 80: 7, 100: 35 & undecided \\
 & 0.6, 0.722 & lottery & 0.93 & 1 & 20: 84, 40: 12, 60: 1, 80: 1, 100: 0 & collapses \\
 & 0.6, 0.737 & checked & 0.95 & 1 & 20: 88, 40: 47, 60: 30, 80: 43, 100: 82, 120: 80, 140: 82, 160: 85 & survives \\
 & 0.6, 0.737 & checked & 0.95 & 2 & 20: 89, 40: 61, 60: 42, 80: 6, 100: 38, 120: 81, 140: 75, 160: 68 & recovers \\
 & 0.6, 0.737 & checked & 0.95 & 3 & 20: 90, 40: 54, 60: 48, 80: 71, 100: 62, 120: 86, 140: 89, 160: 78 & survives \\
 & 0.6, 0.737 & checked & 0.95 & 4 & 20: 88, 40: 58, 60: 39, 80: 72, 100: 69, 120: 51, 140: 42, 160: 83 & survives \\
 & 0.6, 0.737 & lottery & 0.95 & 1 & 20: 85, 40: 12, 60: 0, 80: 0, 100: 0, 120: 0, 140: 0, 160: 0 & collapses \\
 & 0.6, 0.737 & lottery & 0.95 & 2 & 20: 86, 40: 41, 60: 21, 80: 2, 100: 1, 120: 1, 140: 1, 160: 1 & collapses \\
 & 0.6, 0.737 & lottery & 0.95 & 3 & 20: 89, 40: 48, 60: 5, 80: 0, 100: 0, 120: 0, 140: 0, 160: 0 & collapses \\
 & 0.6, 0.737 & lottery & 0.95 & 4 & 20: 83, 40: 15, 60: 0, 80: 0, 100: 0, 120: 0, 140: 0, 160: 0 & collapses \\
 & 0.6, 0.831 & checked & 1.07 & 1 & 20: 89, 40: 84, 60: 90, 80: 96, 100: 94 & survives \\
 & 0.6, 0.831 & checked & 1.07 & 2 & 20: 93, 40: 94, 60: 96, 80: 97, 100: 96 & survives \\
 & 0.6, 0.831 & lottery & 1.07 & 1 & 20: 88, 40: 90, 60: 93, 80: 96, 100: 87 & survives \\
 & 0.6, 0.97 & checked & 1.25 & 1 & 20: 94, 40: 96, 60: 98, 80: 99, 100: 96 & survives \\
 & 0.6, 0.97 & checked & 1.25 & 2 & 20: 91, 40: 97, 60: 99, 80: 98, 100: 98 & survives \\
 & 0.6, 1.242 & checked & 1.60 & 1 & 20: 97, 40: 98, 60: 99, 80: 99, 100: 99 & survives \\
 & 0.6, 1.553 & checked & 2.00 & 1 & 20: 98, 40: 98, 60: 99, 80: 99, 100: 99 & survives \\
 & 0.6, 2.329 & checked & 3.00 & 1 & 20: 99, 40: 99, 60: 99, 80: 99, 100: 100 & survives \\
 & 0.6, 2.329 & lottery & 3.00 & 1 & 20: 98, 40: 100, 60: 99, 80: 100, 100: 100 & survives \\
\midrule
depth 2, sharpened, $p$=0.86 & 0.6, 0.419 & checked & 0.60 & 1 & 20: 97, 40: 34, 60: 0, 80: 0, 100: 0 & collapses \\
 & 0.6, 0.628 & checked & 0.90 & 1 & 20: 98, 40: 95, 60: 84, 80: 62, 100: 49 & undecided \\
 & 0.6, 0.768 & checked & 1.10 & 1 & 20: 99, 40: 98, 60: 96, 80: 94, 100: 95 & survives \\
 & 0.6, 1.117 & checked & 1.60 & 1 & 20: 99, 40: 97, 60: 98, 80: 99, 100: 99 & survives \\
\midrule
depth 2, after a generous phase & 0.6, 0.737 & checked, from lottery & 0.93 & 1 & 20: 97, 40: 83, 60: 67, 80: 29, 100: 83 & survives \\
 & 0.6, 0.737 & checked, from checked & 1.06 & 1 & 20: 97, 40: 89, 60: 94, 80: 98, 100: 99 & survives \\
\midrule
depth 3, $p$=0.66 & 0.6, 0.453 & checked & 0.50 & 1 & 20: 2, 40: 0, 60: 0, 80: 0, 100: 0 & collapses \\
 & 0.6, 0.725 & checked & 0.80 & 1 & 20: 29, 40: 1, 60: 0, 80: 0, 100: 0 & collapses \\
 & 0.6, 0.831 & checked & 0.92 & 1 & 20: 37, 40: 8, 60: 18, 80: 25, 100: 36, 120: 10, 140: 16, 160: 20 & collapses \\
 & 0.6, 0.831 & checked & 0.92 & 2 & 20: 51, 40: 15, 60: 18, 80: 29, 100: 37, 120: 34, 140: 41, 160: 47 & undecided \\
 & 0.6, 0.831 & checked & 0.92 & 3 & 20: 46, 40: 4, 60: 4, 80: 6, 100: 22, 120: 12, 140: 16, 160: 36 & undecided \\
 & 0.6, 0.831 & lottery & 0.92 & 1 & 20: 49, 40: 3, 60: 1, 80: 0, 100: 0, 120: 0, 140: 0, 160: 0 & collapses \\
 & 0.6, 0.831 & lottery & 0.92 & 2 & 20: 51, 40: 1, 60: 1, 80: 1, 100: 1, 120: 1, 140: 0, 160: 0 & collapses \\
 & 0.6, 0.831 & lottery & 0.92 & 3 & 20: 48, 40: 1, 60: 1, 80: 0, 100: 0, 120: 0, 140: 0, 160: 0 & collapses \\
 & 0.6, 0.97 & checked & 1.07 & 1 & 20: 57, 40: 59, 60: 79, 80: 78, 100: 71 & survives \\
 & 0.6, 0.97 & checked & 1.07 & 2 & 20: 67, 40: 56, 60: 76, 80: 69, 100: 77 & survives \\
 & 0.6, 1.132 & checked & 1.25 & 1 & 20: 74, 40: 83, 60: 89, 80: 83, 100: 88 & survives \\
 & 0.6, 1.812 & checked & 2.00 & 1 & 20: 86, 40: 93, 60: 92, 80: 93, 100: 93 & survives \\
\midrule
depth 3, sharpened, $p$=0.77 & 0.6, 0.725 & checked & 0.93 & 1 & 20: 79, 40: 19, 60: 20, 80: 18, 100: 38 & undecided \\
 & 0.6, 0.831 & checked & 1.07 & 1 & 20: 84, 40: 82, 60: 75, 80: 74, 100: 88, 120: 85, 140: 70, 160: 78 & survives \\
 & 0.6, 0.831 & checked & 1.07 & 2 & 20: 87, 40: 67, 60: 69, 80: 85, 100: 91, 120: 84, 140: 76, 160: 82 & survives \\
 & 0.6, 0.831 & checked & 1.07 & 3 & 20: 88, 40: 80, 60: 86, 80: 84, 100: 83, 120: 85, 140: 86, 160: 82 & survives \\
\end{longtable}
}

\begin{table}[htbp]
\caption{Every code run ($b=0.3$, seed 1).
For each run we give BUGGY recall on the 337 held-out buggy programs at each evaluation.
At the last evaluation we also give the share of the 337 correct programs (copies of the 181 references) called CORRECT, and the validity of sampled counterexample attempts.
In the coder's three runs that collapse, this validity rests on 3 to 5 attempts.
For the bases these three numbers are 66, 80, and 73\% (3B coder) and 65, 74, and 71\% (Phi-3.5-mini).
$pr/b$ uses each base's own rate.
The buggy-only run trains on buggy programs alone, which sets $\kappa=0$ (\cref{app:shared}).}
\label{tab:code}
\centering\scriptsize
\setlength{\tabcolsep}{3pt}
\begin{tabular}{@{}lccp{7.4cm}ccl@{}}
\toprule
model & $r$ & $pr/b$ & BUGGY recall (\%) at each evaluation (step: value) & \shortstack{correct called\\correct (\%)} & \shortstack{validity of\\attempts (\%)} & outcome \\
\midrule
3B coder & 0.204 & 0.50 & 10: 19, 20: 0, 30: 0 & 100 & 60 & collapses \\
 & 0.327 & 0.80 & 10: 20, 20: 0, 30: 0 & 100 & 33 & collapses \\
 & 0.388 & 0.95 & 10: 22, 20: 0, 30: 0, 40: 0, 60: 0, 80: 0, 100: 0 & 100 & 75 & collapses \\
 & 0.51 & 1.25 & 10: 26, 20: 1, 30: 0, 40: 0, 60: 1, 80: 32, 100: 39 & 97 & 81 & undecided \\
 & 0.817 & 2.00 & 10: 51, 20: 50, 30: 70, 40: 82, 60: 80, 80: 81, 100: 74 & 79 & 80 & survives \\
 & 1.225 & 3.00 & 10: 87, 20: 97, 30: 100, 40: 99, 60: 90, 80: 90, 100: 93 & 45 & 76 & survives \\
\quad buggy only & 0.51 & 1.25 & 10: 92, 20: 100, 30: 100 & 0 & 73 & survives \\
\midrule
Phi-3.5-mini & 0.339 & 0.80 & 10: 39, 20: 4, 30: 1 & 100 & 86 & collapses \\
 & 0.403 & 0.95 & 10: 39, 20: 6, 30: 2, 40: 1, 60: 1, 80: 2, 100: 3 & 100 & 85 & collapses \\
 & 0.53 & 1.25 & 10: 42, 20: 13, 30: 7, 40: 7, 60: 15, 80: 26, 100: 37, 120: 36, 140: 48, 160: 39, 180: 41, 200: 49 & 94 & 80 & undecided \\
 & 0.678 & 1.60 & 10: 45, 20: 24, 30: 24, 40: 32, 60: 47, 80: 51, 100: 58 & 86 & 79 & undecided \\
 & 0.848 & 2.00 & 10: 55, 20: 47, 30: 48, 40: 56, 60: 62, 80: 72, 100: 66 & 78 & 77 & survives \\
 & 1.272 & 3.00 & 10: 72, 20: 89, 30: 97, 40: 98, 60: 98, 80: 93, 100: 87 & 52 & 74 & survives \\
\bottomrule
\end{tabular}

\end{table}

\FloatBarrier
\section{Training details}
\label{app:training}

\paragraph{Software and hardware.} We used verl 0.7.1, with vLLM 0.11.0 for rollouts, PyTorch 2.8.0, and transformers 4.57.6, and we used TRL for supervised fine-tuning (SFT).
Each run used one GPU.
Our Datalog and code runs used an RTX PRO 6000 (96GB), and a Datalog run took about 10 hours per 100 steps.
The abstention runs used an RTX 6000 Ada (48GB).
The runs on a second cluster used an A100 80GB, with the same code and data, checked by checksum.
In all, the RL runs reported here took about 810 GPU-hours of training, about 410 of them on our GPUs and about 400 on the second cluster.
Code, configurations, and result files will be released with a future version (\cref{app:tables}).

\paragraph{Datalog data and SFT bases.}
Instances have 20 atoms and rule bodies of size 2, and they come in the twin pairs of \cref{sec:setup}.
Both bases are Qwen2.5-3B-Instruct, fine-tuned on the same 11{,}700 instances with checker-valid proofs and closures.
The fine-tuning used learning rate $2\times10^{-5}$, batch 4 with gradient accumulation 8, and 8-bit AdamW, for three epochs for the depth-2 base and five for the depth-3 base.
The RL training sets have 4{,}743 instances whose proofs are two levels deep and 4{,}957 whose proofs are three levels deep.
They are deduplicated against the fine-tuning data and every evaluation set, up to renaming and rule order.
Honesty is measured on the 400 underivable instances of a held-out set of 800, and the probes of \cref{app:frozen} are fit and tested on the same 800.
The rate $p$ of each Datalog base is measured at temperature 1 on 2,000 fresh underivable instances, with 8 samples each.
For honesty, verdicts and certificates are generated greedily with vLLM.

\paragraph{GRPO.}
We train all parameters on one GPU, with FSDP2, bf16 mixed precision, gradient checkpointing, and the parameters and optimizer state offloaded to host memory.
The learning rate is $5\times 10^{-7}$.
Each step uses 64 prompts in one PPO mini-batch, with micro-batch 2.
Each prompt gets 8 rollouts at temperature 1.0 and top-$p$ 1.0.
Prompts are at most 1024 tokens long and responses at most 768.
The loss has a KL term to the reference model with coefficient 0.001 (low-variance estimator), no entropy bonus, and token-mean aggregation.
Group advantages are centered on the group mean without division by the group standard deviation.
This is the advantage form of \citet{liu2025drgrpo}, and \cref{app:std} discusses the form with the division.
Datalog runs last 100 or 160 steps, with a checkpoint every 20 steps (every 10 for one seed at $pr/b=0.80$).
Seeds change the data order.

\paragraph{Rewards.}
A Datalog reward is a table over the checker's outcome cells $\{$valid proof, DERIVABLE claim with invalid or missing proof, valid refutation, invalid refutation$\}$.
The sharpened bases were made with the symmetric soundness reward $\{1, 0, 1, 0\}$, and the runs of the main text use $\{1, b, r, 0\}$.
The sharpening RL runs for 100 steps at depth 2, and for two rounds of 100 at depth 3.

\paragraph{Code task.}
The coder base was fine-tuned from Qwen2.5-Coder-3B-Instruct on the 1,059 balanced examples of \cref{app:code}, with the script used for the 3B Datalog base.
It was trained for three epochs at learning rate $2\times10^{-5}$, with effective batch 32, 8-bit AdamW, and maximum sequence length 1536.
Phi-3.5-mini-instruct was fine-tuned the same way, with maximum sequence length 1024.
GRPO on code uses the Datalog settings, except that the micro-batch is 4 and the parameters stay on the GPU, on one RTX PRO 6000.
Runs are evaluated at steps 10, 20, 30, 40, 60, 80, and 100, and the 200-step run every 20 steps after that.
Each evaluation decodes greedily and also draws 4 samples per program (8 for the bases).
The rollout engine's context is capped at 2,048 tokens in every Datalog and code run.

\paragraph{The lottery.}
The lottery is a switch in the reward function.
On an underivable instance it pays an honest verdict $r$ with probability $\ell$ and 0 otherwise, drawn independently for each rollout, without calling the checker.
Everything else, including the evaluations, is unchanged.
We set $\ell$ to the base's $p$ (\cref{sec:setup}), which is 0.7729 at depth 2 and 0.6624 at depth 3.

\section{The general reward table}
\label{app:table}

\begin{table}[htbp]
\caption{The general reward table, with a payoff for each (truth, verdict, certificate) outcome.
A dash marks an outcome that cannot occur.
A checker sees only the column, so a verifiable reward pays one amount throughout each column ($B{=}D$, $C{=}F$).}
\label{tab:general}
\centering\small
\begin{tabular}{lcccc}
\toprule
& \multicolumn{2}{c}{DERIVABLE} & \multicolumn{2}{c}{UNDERIVABLE} \\
\cmidrule(lr){2-3}\cmidrule(lr){4-5}
instance & valid proof & no valid proof & valid closure & no valid closure \\
\midrule
derivable & $A$ & $B$ & -- & $C$ \\
underivable & -- & $D$ & $E$ & $F$ \\
\bottomrule
\end{tabular}
\end{table}

Give each of the six (truth, verdict, certificate) outcomes its own payoff (\cref{tab:general}).
On a derivable instance, the reward pays $A$ for a valid proof, $B$ for a rejected proof, and $C$ for an UNDERIVABLE verdict.
On an underivable instance, it pays $D$ for a DERIVABLE claim, $E$ for a valid refutation, and $F$ for a rejected one.
A checker cannot tell a rejected proof of a derivable goal from a bluff on an underivable one, or a wrong UNDERIVABLE from a failed refutation, since each pair shares a column of the table.
So a verifiable reward has $B{=}D$ and $C{=}F$.
Any other reward needs the truth label.
Let $p$ be the rate at which refutations are valid and $q$ the rate at which proofs are valid.
Then the argument of \cref{sec:threshold} gives one condition for each side:
\begin{equation}
\text{refutations survive} \iff p\,(E-F) > D-F, \qquad \text{proofs survive} \iff q\,(A-B) > C-B .
\end{equation}
Adding a constant to all six payoffs cancels in the group mean, and multiplying them by a positive factor preserves every sign.
So only these two combinations decide the outcome.

\begin{proposition}[One side at most]
\label{prop:oneside}
Let a reward be verifiable ($B=D$, $C=F$) and pay a valid certificate more than a rejected one ($A>B$, $E>F$), and let $p,q>0$.
Then the two conditions above cannot both fail.
In other words, a bluff is the sure thing on at most one side.
\end{proposition}
\begin{proof}
Since $p(E-F)>0$, refutations can fail only if $D-F>0$, and since $q(A-B)>0$, proofs can fail only if $C-B>0$.
With $B=D$ and $C=F$ these say $D>F$ and $F>D$, which cannot both hold.
\end{proof}
Our reward is $A{=}1$, $B{=}D{=}b$, $C{=}F{=}0$, $E{=}r$, and under it proofs always survive.

\section{The code task}
\label{app:code}

\paragraph{Data.}
From all 974 MBPP problems we generate single-site mutants of the reference solution, i.e.\ copies changed in one place (comparison, arithmetic, and boolean operator swaps; off-by-one changes to constants, ranges, and indices; dropped conditions).
We keep a mutant only if it fails at least one hidden test while the reference passes all of them.
This leaves 7,358 real bugs over 726 problems.
The RL set has 11,076 prompts over 545 problems, half buggy and half correct (each correct program is repeated to match its mutants).
The held-out set has 674 prompts over 181 problems.
It has up to two mutants per problem (337 in all), each paired with a copy of the reference.

\paragraph{Oracle and reward.}
A counterexample is one call to the function with literal positional arguments.
It is valid if the reference runs on it without error and the function under test returns a different value, raises an exception, or times out.
Both functions run in fresh isolated subprocesses.
The reward pays $b=0.3$ for CORRECT on any program, $r$ for a valid counterexample on a buggy program, and 0 otherwise, including any counterexample on a correct program.

\paragraph{Bases.}
Before fine-tuning, Qwen2.5-Coder-3B-Instruct calls 62\% of held-out buggy programs buggy and 71\% of correct ones correct, and its sampled counterexamples are valid 61\% of the time.
We fine-tune it for three epochs on 1,059 examples from 545 training problems.
Each problem contributes one correct program.
For the 514 problems where a failing hidden test gives a usable call, it also contributes one buggy program, with a counterexample read off that test.
The correct programs are there so that the model is not taught to call everything buggy.
The fine-tuned coder calls 66\% of buggy programs buggy and 80\% of correct ones correct, with $p=0.735$.
Phi-3.5-mini-instruct, fine-tuned the same way, calls 65\% of buggy programs buggy and 74\% of correct ones correct, with $p=0.708$.

\section{Abstention}
\label{app:abstention}

\paragraph{Setup.}
The base is Qwen2.5-1.5B-Instruct without fine-tuning.
GSM8K is a set of grade-school math word problems.
For each problem, the prompt asks for a final line \texttt{ANSWER: <number>}, or \texttt{ANSWER: I don't know} if the model is not sure.
The reward is 1 for a correct number, 0 for ``I don't know'', and $-\lambda$ for a wrong or unparseable answer.
Training draws from the 7,473 problems of the GSM8K training split, with the GRPO settings of \cref{app:training}.
On 2,000 of these problems the base's sampled accuracy when it answers is $p=0.701$, from 8 samples each.
Every 20 steps we evaluate on 500 held-out problems, greedily and with 4 samples.
For the per-problem rates below, the base was sampled 8 times on each held-out problem.

\paragraph{Results.}
The penalty at the threshold is $\lambda^*=p/(1-p)=2.34$.
We trained the base for 100 steps at $\lambda^*/4$ and $4\lambda^*$, with two seeds each, and at $0.8\lambda^*$ and $1.25\lambda^*$, with one seed each.
Below $\lambda^*$ the model keeps answering.
At step 100 the runs at $\lambda^*/4$ abstain on 1\% (seed 1) and 3\% (seed 2) of 500 held-out problems, and the run at $0.8\lambda^*$ on 3\%, against 2\% before RL.
At $4\lambda^*$ both seeds abstain on 99\% or more, including every problem the base solved in all eight samples.
The run at $1.25\lambda^*$ ($pr/b=0.94$) still answers 94\%.
At this penalty, answering pays more than abstaining once the model's accuracy on the problems it answers passes $\lambda/(1+\lambda)=0.75$.
The run keeps answering because its accuracy when it answers the training problems rises past that value within 40 steps, as in a run trained with the checker below the threshold (\cref{sec:race}).
At $4\lambda^*$ the model's accuracy when it answers reaches 0.80 in seed 1 and 0.77 in seed 2, short of the 0.90 that answering needs at this penalty.
In the training samples its answers die out between steps 40 and 50 in seed 2 and between steps 40 and 80 in seed 1.

\paragraph{The rise is improvement.}
At $1.25\lambda^*$, the accuracy when answering could rise just because the model abstains on hard problems.
At $\lambda^*/4$, where the model hardly abstains, accuracy when answering also rises, from 0.71 to 0.79 by step 40 in seed 1, and from 0.73 to 0.76 by step 40 and 0.79 by step 60 in seed 2.

\paragraph{Where it abstains.}
At $1.25\lambda^*$ the model abstains on 11\% of the held-out problems it solved in at most five of eight samples before RL, and on 1\% of the rest.
Whether the model abstains on a problem depends on more than that problem's own rate.
Below $\lambda^*$, most problems whose rate before RL is below $\lambda/(1+\lambda)$ are still answered, and at $4\lambda^*$ the model abstains even on problems it always solved.

\section{A shared verdict}
\label{app:shared}

The comparison of \cref{sec:threshold} looks at one class only.
In this appendix we derive the threshold when the answer BUGGY or UNDERIVABLE can also be given on the other class (correct programs, derivable instances), where it is wrong.
We write the derivation for code.
The Datalog case differs only in one payoff, which we give in the paragraph on Datalog below.

\paragraph{Setup.}
On code, the class at stake, $t$ (for target), is the set of buggy programs, and the other class, $o$, is the set of correct programs.
Consider a program $i$, let $h_i$ be the probability that the model answers BUGGY on it, and let
\[
x_i=\log\frac{h_i}{1-h_i}
\]
be its log-odds, which is a function of the weights $w$.
Inverting this definition gives $h_i=\sigma(x_i)$, where $\sigma(x)=1/(1+e^{-x})$ is the logistic function.
Since the fine-tuned bases begin their answers with the verdict, $x_i$ is the difference between the log-probabilities of the two verdicts right after the prompt.
On a buggy program, CORRECT pays $b$, and BUGGY pays $r$ if its counterexample is valid and 0 otherwise.
The counterexample is valid with probability $p_i$.
On a correct program, CORRECT pays $b$ and BUGGY pays 0, since no counterexample can be valid.
For each program in the batch, GRPO samples a group of $n$ answers $a=1,\dots,n$.
It gives each answer the advantage $A_a=R_a-\bar R$, where $R_a$ is the answer's reward and $\bar R$ is the group's mean reward.
It then takes a gradient step that raises each answer's log-probability $\log\pi(a)$ in proportion to $A_a$.

We now prove \cref{prop:step}, which we restate.

\propstep*

The proof rests on two lemmas.
The first says what one group does to the verdict, and the second says how one step carries the effect of the groups in a batch to every program.
The argument applies the first-order account of how an update on one example moves another \citep{ren2025learning} to GRPO's group advantages and to a verdict that the two classes share.

\begin{lemma}[One group]
\label{lem:group}
Let program $i$'s group have $n$ independently sampled answers with advantages $A_a=R_a-\bar R$, and let $B_a=1$ if answer $a$ says BUGGY and $B_a=0$ otherwise.
The parts of the group's updates that come from the verdict's log-probability add up to
\[
\sum_a A_a\,\nabla_w\log\pi(\textup{verdict of }a)=P_i\,\nabla_w x_i, \qquad P_i=\sum_a A_aB_a,
\]
the total advantage of the group's BUGGY answers, and
\begin{equation}
\mathbb{E}P_i=(n-1)\,h_i(1-h_i)\,(\mu_B-\mu_C),
\label{eq:EP}
\end{equation}
where $\mu_B$ and $\mu_C$ are the expected rewards of a BUGGY and of a CORRECT answer on program $i$.
On a buggy program $\mu_B-\mu_C=p_ir-b$, and on a correct one it is $-b$.
\end{lemma}

\begin{proof}
We first show that every answer's verdict gradient points in the same direction, and then take expectations.
Split $\log\pi(a)$ into the log-probability of the verdict and the log-probability of the rest of the answer given the verdict.
The verdict's log-probability is $\log h_i$ if the verdict is BUGGY and $\log(1-h_i)$ if it is CORRECT, and since $h=\sigma(x)$, the derivatives of these two with respect to the log-odds are
\[
\frac{d}{dx}\log h=\frac{d}{dx}\bigl(-\log(1+e^{-x})\bigr)=\frac{e^{-x}}{1+e^{-x}}=1-h,
\qquad
\frac{d}{dx}\log(1-h)=\frac{d}{dx}\bigl(-x-\log(1+e^{-x})\bigr)=-1+(1-h)=-h.
\]
In other words, both cases are covered by the single formula
\begin{equation}
\nabla_w\log\pi(\textup{verdict of }a)=(B_a-h_i)\,\nabla_w x_i ,
\label{eq:score}
\end{equation}
and so every answer's verdict gradient points along the same direction, $\nabla_w x_i$.
Summing over the group gives
\[
\sum_a A_a(B_a-h_i)\,\nabla_w x_i=\Bigl[\sum_a A_aB_a-h_i\sum_a A_a\Bigr]\nabla_w x_i=P_i\,\nabla_w x_i,
\]
since the advantages are deviations from the group mean and so sum to zero.
It remains to compute the expectation.
For this, let $m_i=h_i\mu_B+(1-h_i)\mu_C$ be the expected reward of any one answer.
By linearity of expectation, and since $B_a$ is 0 or 1,
\[
\mathbb{E}P_i=\sum_a\mathbb{E}[A_aB_a]=\sum_a h_i\,\mathbb{E}[A_a\mid B_a=1].
\]
To compute the conditional expectation, write $A_a=\frac{n-1}{n}R_a-\frac1n\sum_{a'\ne a}R_{a'}$.
Given $B_a=1$, the reward $R_a$ has mean $\mu_B$, while each other answer is sampled independently of answer $a$, so its reward has mean $m_i$.
Therefore $\mathbb{E}[A_a\mid B_a=1]=\frac{n-1}{n}(\mu_B-m_i)=\frac{n-1}{n}(1-h_i)(\mu_B-\mu_C)$, and summing over the $n$ answers gives \cref{eq:EP}.
The two values of the gap follow from the payoffs.
On a buggy program, BUGGY earns $r$ with probability $p_i$ and CORRECT earns $b$.
On a correct program, BUGGY earns 0 and CORRECT earns $b$.
\end{proof}

In other words, a program's group pushes the verdict (i.e.\ moves its log-odds in expectation) in the direction of the payoff gap between the two verdicts, and it does so only while the group is likely to hold both verdicts.
The factor $h_i(1-h_i)$ measures how likely that is.
The rest of each answer, the counterexample, contributes other terms to the update.
These terms change which counterexamples the model writes, and so change $p_i$ (\cref{sec:race}).
We set them aside here.

\begin{lemma}[One step]
\label{lem:step}
A step of gradient ascent at step size $\eta$ adds up the updates from the groups of the programs $i$ in its batch, so by \cref{lem:group} the verdict part of the update is $\Delta w=\eta\sum_i P_i\,\nabla_w x_i$.
To first order in the step, it changes the log-odds on any program $j$, in the batch or not, by
\[
\Delta x_j\approx\eta\sum_i P_i\,\langle\nabla_w x_i,\nabla_w x_j\rangle,
\qquad\text{so}\qquad
\mathbb{E}\,\Delta x_j\approx\eta\sum_i\mathbb{E}P_i\,\langle\nabla_w x_i,\nabla_w x_j\rangle .
\]
\end{lemma}

\begin{proof}
To first order, $x_j(w+\Delta w)\approx x_j(w)+\nabla_w x_j\cdot\Delta w$.
Substituting $\Delta w$ and using the linearity of the inner product gives the first formula.
The gradients depend on the weights and the programs but not on which answers were sampled, so the expectation passes through them, and the second formula follows.
\end{proof}

So the inner product $\langle\nabla_w x_i,\nabla_w x_j\rangle$ measures how far an update computed on program $i$ moves the verdict on program $j$ \citep{jacot2018ntk,ren2025learning}.
RL updates, measured on LoRA adapters, can be nearly orthogonal across tasks \citep{zhu2026coexists}.
Our inner products, however, concern one readout that both classes share (the verdict's log-odds), and we measure them below.
Low gradient similarity to the other examples also marks the examples that RLVR fails to learn \citep{chen2026unlearnability}.

\begin{proof}[Proof of \cref{prop:step}]
We average the inner products by kind of pair and then collect the terms from each class.
Let $K$ be the average of $\langle\nabla_w x_i,\nabla_w x_j\rangle$ over pairs of distinct buggy programs, and let $sK$ be its average over pairs of one buggy and one correct program, so that $s$ is the ratio in the proposition.
As in the proposition, we assume $K>0$ and $s>0$.
Replace each inner product in \cref{lem:step} by the average for its kind of pair, and insert \cref{eq:EP}.
For a buggy program $j$ outside the batch, this gives
\[
\mathbb{E}\,\Delta x_j\approx\eta K(n-1)\Bigl[\sum_{i\text{ buggy}}h_i(1-h_i)(p_ir-b)\;-\;s\sum_{i\text{ correct}}h_i(1-h_i)\,b\Bigr].
\]
To read off the sign of the bracket, define the weight of each class and the weighted success rate on the class at stake,
\[
W_t=\sum_{i\text{ buggy}}h_i(1-h_i),\qquad W_o=\sum_{i\text{ correct}}h_i(1-h_i),\qquad \tilde p=\frac1{W_t}\sum_{i\text{ buggy}}h_i(1-h_i)\,p_i ,
\]
which we interpret after the proof.
In these terms the bracket is $W_t(\tilde p r-b)-s\,W_o\,b=W_t\,b\,(\tilde p r/b-1-\kappa)$, with $\kappa=s\,W_o/W_t$.
Since $K>0$, the expected change is therefore positive iff
\[
\frac{\tilde p\,r}{b}>1+\kappa .
\]
On Datalog only the payoff gap on the other class changes, from $-b$ to $-c$ (see below), and the same computation gives $\kappa=s\,W_oc/(W_tb)$.
\end{proof}

Replacing each inner product by its class average is the one modeling assumption of \cref{prop:step}.
This assumption holds in expectation if the inner products vary from pair to pair independently of how hard each program pushes.
The right-hand side of the formula for the expected change is the same for every buggy program, including the held-out ones on which honesty is measured, which have no group of their own.
A training program that is in the batch also gets the term from its own group, which is one of the 64 groups in the batch.
With $s=0$ or $W_o=0$, the condition reduces to the comparison of \cref{sec:threshold}.
The condition gives the direction of a step at the current $h$ and $\tilde p$.

\paragraph{How to read $W_t$, $W_o$, $\tilde p$, $s$, and $\kappa$.}
\begin{itemize}\setlength{\itemsep}{1pt}\setlength{\parsep}{0pt}\setlength{\topsep}{2pt}
\item $W_t$ and $W_o$ count the programs of each class that are still in play.
A program whose verdict is a coin flip adds $1/4$, and one that always gets the same verdict adds 0.
Their ratio $W_o/W_t$ is the weight of the other class relative to the class at stake, before the payoff gaps and the coupling enter.
\item $\tilde p$ is the success rate of the attempts that still count, since programs the model always or never calls buggy carry no weight in it.
On the depth-2 Datalog base it is close to the plain rate $p$ (0.78 against 0.77, from 16 samples on each of 1,000 training instances weighted by the teacher-forced $h(1-h)$).
\item $s$ is how far a push on the correct programs moves the verdict on the buggy ones, relative to how far a push on the buggy programs moves it.
It is 0 if the model moves its verdicts on the two classes independently, and 1 if it treats every program alike.
The coupling is asymmetric.
The coupling in the other direction, $s'$, divides by the inner products within the correct programs instead (\cref{prop:settle}).
\item $\kappa$ is the extra margin that $\tilde p\,r/b$ needs for honesty to rise from the first step.
It is the other class's push, scaled by the coupling and measured in units of the push that each unit of $\tilde p\,r/b-1$ gives the class at stake.
It falls as the correct programs settle and $W_o$ shrinks.
\end{itemize}

\paragraph{What is approximate.}
We treat the verdict as one binary choice made at the start of the answer, and we ignore answers that give neither verdict.
We expand the step to first order.
Over many steps, though, the gradients themselves change.
Training uses Adam, which scales each weight's step by roughly the inverse square root of that weight's mean squared gradient, so that $\Delta w\approx\eta Dg$ with $D$ a diagonal matrix applied to the gradient.
The inner products are then weighted by $D$, and below we measure $s$ both with and without this weighting.
The counterexample parts of the updates, set aside after \cref{lem:group}, also move $x_j$ a little through their own inner products with $\nabla_w x_j$.
The loss averages over all tokens of the batch.
This rescales every term by the same constant, which we absorb into $\eta$.

\paragraph{Datalog.}
On Datalog the class at stake is the set of underivable instances, with UNDERIVABLE in place of BUGGY, and the DERIVABLE claim pays $b$, as CORRECT does on code.
On a derivable instance, UNDERIVABLE is wrong and pays 0, while DERIVABLE pays 1 with a valid proof and $b$ without one.
So on a derivable instance $i$ the gap in \cref{lem:group} is $-c_i$ instead of $-b$, where $c_i=q_i+(1-q_i)b$ is the expected payoff of DERIVABLE given its rate $q_i$ of valid proofs.
In \cref{prop:step}, $c$ is the average of the $c_i$ weighted by $h_i(1-h_i)$, and it is about 1.
On Datalog the derivable class settles within about 13 to 18 steps, so $W_o$, and with it $\kappa$, falls close to 0 (\cref{app:finite}).
On code the correct programs settle too, within about 20 steps at $pr/b=1.25$ in the held-out samples, and the next paragraph measures $\kappa$ at the starting checkpoint.

\paragraph{On code.}
We sample eight answers on each of the 337 buggy held-out programs and the 337 correct ones (copies of 181 references).
On these samples, $W_o/W_t$ is 1.04 for the coder before RL and 0.99 for Phi.
Weighting the success rates by $h(1-h)$ gives $\tilde p=0.67$ for the coder, against the plain rate $p=0.735$, and 0.67 against 0.708 for Phi.
We estimate $s$ on the same programs at each model's starting checkpoint.
Here $x$ is the teacher-forced log-odds of the two verdicts, i.e.\ it is computed from the probability that the model assigns to each verdict placed directly after the prompt, without sampling.
Split the buggy programs at random into two halves, and let $G_t$ and $G'_t$ be the mean gradients of $x$ over the two halves, weighted by $h(1-h)$, with $h$ the share of BUGGY among the eight samples.
Let $G_o$ be the same weighted mean over the correct programs.
We set $s=\tfrac12\bigl(\langle G_t,G_o\rangle+\langle G'_t,G_o\rangle\bigr)/\langle G_t,G'_t\rangle$, so that no gradient is paired with itself.
Since training uses Adam, we also compute the inner products weighted by the inverse square root of the mean squared gradient, as Adam's update sees them.
For the coder, $s=0.89$ (0.84 with Adam's weighting, jackknife standard error 0.02), so $\kappa=0.92$ (0.88) and $1+\kappa=1.92$ (1.88).
For Phi, $s=0.99$ under both weightings (standard error 0.01), so $\kappa=0.98$ (0.97) and $1+\kappa=1.98$ (1.97).
On the depth-2 Datalog base we measured $s$ in the same way, with underivable instances in place of buggy programs and with teacher-forced weights.
We drew 1,000 underivable and 1,000 derivable instances from the RL training set and weighted each by $h(1-h)$, the expected weight of its group under GRPO, where $h$ is the teacher-forced probability of UNDERIVABLE.
The base is nearly unanimous on most instances, so we kept the 803 instances (330 underivable, 473 derivable) with $h(1-h)\ge10^{-4}$, which carry 99.8\% of the weight.
This gives $s=1.03$ (1.02 with Adam's weighting, jackknife standard error 0.07).
On the depth-3 base, which is less settled, the same procedure keeps 1,621 instances (663 underivable, 958 derivable) and gives $s=0.98$ (0.97, standard error 0.02).
The code values of $1+\kappa$ above are in units of $\tilde p$, while our tables and figures use $p$.
In the units of the tables and figures, the expected step on the class at stake therefore changes sign at $1+\kappa$ times the ratio of the plain rate to the weighted one, which is $pr/b\approx2.1$ for both models.

\paragraph{The other class settles below $1+\kappa$.}
\begin{proposition}[The other class settles]
\label{prop:settle}
Under the assumptions of \cref{prop:step}, let the gradient inner product be $K_t$ within the class at stake, $K_o>0$ within the other class, and $K_{to}$ across, so $s=K_{to}/K_t$, and let $s'=K_{to}/K_o$.
If $ss'<1$ and $1<\tilde p\,r/b<1+\kappa$, the expected change in $x$ is negative for every item of the other class outside the batch, i.e.\ that class moves toward its right answer.
When the $K$'s are the inner products of the two classes' mean gradients, $ss'\le1$, with equality only if those gradients are parallel.
\end{proposition}
\begin{proof}
By \cref{lem:step} with class averages, the expected change in $x$ on an item of the other class outside the batch is $\eta(n-1)K_o\bigl(s'W_t(\tilde p\,r-b)-W_oc\bigr)$, so since $K_o>0$ it has the sign of $s'W_t(\tilde p\,r-b)-W_oc$.
Suppose $1<\tilde p\,r/b<1+\kappa$, so that $W_o>0$.
Then $W_t(\tilde p\,r-b)$ is positive, and by the definition of $\kappa$ it is below $sW_oc$.
Since $s>0$ and $K_o>0$, also $s'=sK_t/K_o>0$, so $s'W_t(\tilde p\,r-b)-W_oc$ is below $(ss'-1)W_oc$, which is negative because $ss'<1$.
For the last claim, if $K_t=\lvert G_t\rvert^2$, $K_o=\lvert G_o\rvert^2$, and $K_{to}=\langle G_t,G_o\rangle$, then $ss'=\langle G_t,G_o\rangle^2/(\lvert G_t\rvert^2\lvert G_o\rvert^2)$ is the squared cosine between $G_t$ and $G_o$, which is at most 1 by the Cauchy--Schwarz inequality, with equality only if they are parallel.
\end{proof}
In fact, the same computation shows that the push on the other class is negative exactly when $\tilde p\,r/b<1+W_oc/(s'W_tb)=1+\kappa/(ss')$, which lies above $1+\kappa$ when $ss'<1$.
Swapping the classes in the measurement above gives the coupling $s'$ of an update driven by buggy programs on correct ones.
It is 1.08 (1.09 with Adam's weighting) for the coder, 0.93 (0.89) for Phi, 0.86 (0.83) on the depth-2 Datalog base, and 0.95 (0.94) on the depth-3 base.
On our two code models the squared cosine between the mean gradients is 0.91--0.95, and $ss'$ itself is 0.93--0.95 (0.88--0.92 with Adam's weighting).
On the depth-2 and depth-3 Datalog bases $ss'$ is 0.88 and 0.93.
On the depth-2 base the teacher-forced weights give $W_o/W_t=1.57$, so with $c\approx1$ and $b=0.6$, $1+\kappa$ is about 3.7 and the ceiling $1+\kappa/(ss')$ is about 4.0.
The other class therefore settles, and $W_o$ and $\kappa$ fall with it.
Once they have, the comparison of $p_t\,r/b$ with 1 decides where the run ends, provided that its attempts last.
Above $\tilde p\,r/b=1+W_oc/(s'W_tb)$, the push on the other class turns positive, and the model starts calling correct programs buggy.
With the measured couplings this point is at 1.97 for the coder and 2.06 for Phi in units of the weighted rate, or about 2.15--2.2 on the plain rate.
The runs agree with this.
At $pr/b=2$ both models call correct programs correct about as often at step 100 as before RL (79\% and 78\% of the time, against 80\% and 74\%), though the coder dips to 65\% at step 40.
At $pr/b=3$ they do so only 45\% and 52\% of the time (\cref{tab:codecheck}).
At $pr/b=3$ the share falls to 13\% by step 30 for the coder and to 30\% by step 40 for Phi.
Phi's BUGGY recall at $pr/b=2$, just below its $1+\kappa$, first falls from 65\% to 47\% at step 20, then rises to 72\% at step 80, and ends at 66\% at step 100.
At 1.6 it falls to 24\% and then rises to 58\%.
At 3, above both points, it rises from the start, to 97\% at step 30.

\paragraph{Without the other class.}
Since $\kappa$ scales with $W_o/W_t$, it is set by the share of correct programs in the RL training set.
We trained the coder at $pr/b=1.25$ for 30 steps on the buggy programs of its RL set alone, so that $W_o=0$ in every batch and $\kappa=0$.
\Cref{prop:step} then predicts no early dip.
Moreover, the point $1+W_oc/(s'W_tb)$ is then 1, so the correct programs, now absent from training, should drift toward BUGGY.
Both predictions held.
BUGGY recall is 92\%, 100\%, and 100\% at steps 10, 20, and 30, against 26\%, 1\%, and 0\% with the full RL set.
Correct programs are called correct 42\%, 2\%, and 0\% of the time at the same steps (80\% before RL).
With nothing to hold the correct programs in place, the coder ends up answering BUGGY everywhere.
This is why we report both classes on code.
In this run each batch has 64 buggy prompts instead of about 32.
The run has one seed.

\begin{table}[htbp]
\caption{Every code run, placed against $1+\kappa$ (about 2.1 for both models on the plain rate) and against the ratio above which correct programs start being called buggy (about 2.15--2.2), both computed from the coupling measured at the starting checkpoint.
The columns give BUGGY recall (before RL, at the first evaluation, and at the last) and the share of correct programs called correct (before RL and at the last evaluation), in \%.
The row marked buggy only is the run trained on buggy programs alone, which sets $\kappa=0$.}
\label{tab:codecheck}
\centering\small
\begin{tabular}{llcccl}
\toprule
model & $pr/b$ & regime & BUGGY recall & correct called correct & opening \\
\midrule
3B coder & 0.50 & below 1 & 66, 19 (step 10), 0 (step 30) & 80, 100 & falls, collapses \\
 & 0.80 & below 1 & 66, 20 (step 10), 0 (step 30) & 80, 100 & falls, collapses \\
 & 0.95 & below 1 & 66, 22 (step 10), 0 (step 100) & 80, 100 & falls, collapses \\
 & 1.25 & 1 to $1{+}\kappa$ & 66, 26 (step 10), 39 (step 100) & 80, 97 & falls, then rises \\
 & 2.00 & just below $1{+}\kappa$ & 66, 51 (step 10), 74 (step 100) & 80, 79 & falls, then rises \\
 & 3.00 & above 2.2 & 66, 87 (step 10), 93 (step 100) & 80, 45 & rises \\
 & 1.25 & $\kappa=0$ (buggy only) & 66, 92 (step 10), 100 (step 30) & 80, 0 & rises \\
\midrule
Phi-3.5-mini & 0.80 & below 1 & 65, 39 (step 10), 1 (step 30) & 74, 100 & falls, collapses \\
 & 0.95 & below 1 & 65, 39 (step 10), 3 (step 100) & 74, 100 & falls, collapses \\
 & 1.25 & 1 to $1{+}\kappa$ & 65, 42 (step 10), 49 (step 200) & 74, 94 & falls, then rises \\
 & 1.60 & 1 to $1{+}\kappa$ & 65, 45 (step 10), 58 (step 100) & 74, 86 & falls, then rises \\
 & 2.00 & just below $1{+}\kappa$ & 65, 55 (step 10), 66 (step 100) & 74, 78 & falls, then rises \\
 & 3.00 & above 2.2 & 65, 72 (step 10), 87 (step 100) & 74, 52 & rises \\
\bottomrule
\end{tabular}
\end{table}

\section{Standardized advantages}
\label{app:std}

GRPO as introduced \citep{shao2024deepseekmath}, and many implementations by default, divide each group's advantages by the group's standard deviation.
This keeps every sign but weights each group by the inverse of its spread, so instances with success rates near 0 or 1 count for more.
Consider a group of $n$ rollouts (sampled answers) on instance $i$.
Each rollout bluffs with some probability and is then paid $b$, and otherwise it attempts a refutation, which is valid with probability $p_i$ and then paid $r$.
With mean-centered advantages, the expected push toward the bluff on instance $i$ is proportional to $h_i(1-h_i)(b-p_ir)$, where $1-h_i$ is its bluff rate (\cref{lem:group}).
So when every instance has the same bluff rate, the threshold on $r/b$ is $1/\bar p$, where $\bar p$ is the mean of the $p_i$, whatever the group size and that rate.
When the bluff rates differ, it is $1/\tilde p$, with the $p_i$ weighted by $h_i(1-h_i)$ as in \cref{prop:step}.
With standardized advantages, the threshold depends on $n$, on the bluff rate, and on how the $p_i$ spread around their mean.
So two training sets with the same mean $p$ can have different thresholds.
We use mean-centered advantages throughout.
On the depth-2 base we computed both thresholds from per-instance rates.
We sampled 16 answers on each of 1,000 underivable training instances and graded them with the checker.
Of these instances, 995 drew at least one attempt, and their pooled validity is 0.773, the base's $p$.
We then computed the expected push of a group of 8 exactly, by summing over the group's possible compositions.
With mean-centered advantages the push changes sign at $pr/b=0.97$ at the bluff rates of these samples, under which only 61 instances draw a bluff and $\tilde p=0.80$, and at 1.00 for any bluff rate shared by all instances.
With the teacher-forced weights of \cref{app:shared}, which give $\tilde p=0.78$, it would change sign at 0.99.
With standardized advantages it changes sign at $r/b=1$, i.e.\ at $pr/b=0.77$, at the base's own bluff rates and at bluff rates of 0.1, 0.3, and 0.5.
Just above $r/b=1$, a group whose attempts are all valid has rewards that differ only by $r-b$, so dividing by their small spread gives that group a large push toward the attempt.
On this base such groups outweigh the rest.
So with standardized advantages we would expect the threshold to sit near $r=b$, below $pr/b=1$.
We did not train with standardized advantages, so this prediction is untested.
Collapse under the division has been reported for shaped rewards \citep{wang2026darkroom,yu2026phantom}.
\citet{bay2026identity} give the finite-group correction of the divided update for binary rewards.
In a group with two reward values, the standardized advantages depend only on the fraction of answers receiving each value, and for this reason \citet{he2026rlec} center without dividing when abstention rewards encode a confidence target.
Under error-penalized abstention, \citet{che2026abstention} show that when few rollouts answer, the division replaces the designed penalty with an effective penalty of one, which moves the answer threshold to $1/2$.
In a one-step model with two reward channels, one reliably and one unreliably obtained, the group z-score (i.e.\ the standardized advantage) moves GRPO's boundary off the expected-reward line through finite-group effects \citep{li2026multireward}.
The division also reweights prompts by difficulty \citep{liu2025drgrpo} and makes GRPO overconfident on stochastic outcomes \citep{bereket2025uncalibrated}.
Even without the division, group-relative advantages are biased by prompt difficulty within the groups that have mixed outcomes, which are the only groups that produce an update \citep{yang2026biased}.

\section{The frozen probe}
\label{app:frozen}

We compare two probes.
The refit probe is an $\ell_2$-regularized logistic regression on the residual stream at the last prompt token.
It is fit separately for each layer at every checkpoint, on a 70/30 split of 800 instances, and we report its held-out accuracy at the base's chosen layer.
The frozen probe is fit the same way on the base model's activations and then applied unchanged to each trained checkpoint's activations on the held-out split.
We choose the layer by the base's held-out accuracy only.
This gives layer 27 for the depth-2 base, where the probe is 99\% accurate, and layer 29 for the depth-3 base, where it is 90\% accurate.
The copy of the depth-3 base on the second cluster chose layer 28 (90.8\% accurate, against 90.4\% at layer 29 on ours), and we read every depth-3 run at layer 29.
\Cref{tab:frozen} gives, for each run's last checkpoint, the refit probe's accuracy, the frozen probe's accuracy with its original cutoff, and the frozen probe's AUROC (area under the ROC curve), which is independent of the cutoff.
A positive score from the frozen probe reads DERIVABLE.
In the ten runs from the depth-2 base that collapse (seed 1 at $pr/b=0.80$ counted once, by its rerun), the probe's mean score on underivable instances lies between $+2$ and $+13$, against $+22$ to $+27$ on derivable ones.
Its AUROC in these runs is 0.92--0.99 (0.99--1.00 at layer 25).
In the runs that end honest, the underivable mean lies between $-9$ and $-18$, and the AUROC is 0.99--1.00.
At depth 3, where the base's probe is less accurate, the six runs that collapse have AUROC 0.895--0.953.
For runs from the sharpened bases, the probe is the one fit on the unsharpened base of the same depth.

\begin{table}[htbp]
\caption{The base's probe, held fixed, applied to trained checkpoints.
The columns give the underivable recall of the model's output (honest), the refit probe's accuracy, the frozen probe's accuracy with its original cutoff, the frozen probe's AUROC at the layer chosen on the base, its best AUROC over layers, and the layer chosen on the base.
The seed-1 run trained with the checker at $pr/b=0.80$ is the one marked $^\ast$ in \cref{tab:long}.}
\label{tab:frozen}
\centering\scriptsize
\setlength{\tabcolsep}{3pt}
\begin{tabular}{llcccccc}
\toprule
run & seed, step & honest & refit probe & frozen probe & frozen AUROC & frozen AUROC, best layer & base layer \\
\midrule
depth 2, checked, $pr/b=0.40$ & 1, 100 & 0 & 97.5 & 50.8 & 0.916 & 0.990 (layer 25) & 27 \\
depth 2, checked, $pr/b=0.60$ & 1, 100 & 0 & 97.5 & 52.1 & 0.935 & 0.997 (layer 25) & 27 \\
depth 2, checked, $pr/b=0.80$ & 1, 100 & 0 & 97.5 & 54.2 & 0.966 & 0.997 (layer 25) & 27 \\
depth 2, checked, $pr/b=0.80$ & 2, 100 & 0 & 97.5 & 64.2 & 0.976 & 0.995 (layer 25) & 27 \\
depth 2, lottery, $pr/b=0.80$ & 1, 100 & 0 & 97.9 & 52.1 & 0.959 & 0.995 (layer 25) & 27 \\
depth 2, checked, $pr/b=0.93$ & 1, 100 & 76 & 98.3 & 91.7 & 0.998 & 0.999 (layer 25) & 27 \\
depth 2, checked, $pr/b=0.93$ & 2, 100 & 35 & 98.3 & 87.1 & 0.999 & 0.999 (layer 25) & 27 \\
depth 2, lottery, $pr/b=0.93$ & 1, 100 & 0 & 97.9 & 69.2 & 0.987 & 0.997 (layer 25) & 27 \\
depth 2, checked, $pr/b=0.95$ & 1, 160 & 85 & 98.3 & 95.0 & 0.999 & 1.000 (layer 29) & 27 \\
depth 2, checked, $pr/b=0.95$ & 2, 160 & 68 & 97.9 & 91.2 & 0.998 & 0.999 (layer 25) & 27 \\
depth 2, checked, $pr/b=0.95$ & 3, 160 & 78 & 98.3 & 93.3 & 0.998 & 0.998 (layer 28) & 27 \\
depth 2, checked, $pr/b=0.95$ & 4, 160 & 83 & 97.5 & 94.2 & 0.994 & 0.998 (layer 25) & 27 \\
depth 2, lottery, $pr/b=0.95$ & 1, 160 & 0 & 98.8 & 59.6 & 0.981 & 0.997 (layer 25) & 27 \\
depth 2, lottery, $pr/b=0.95$ & 2, 160 & 1 & 97.9 & 72.5 & 0.987 & 0.997 (layer 25) & 27 \\
depth 2, lottery, $pr/b=0.95$ & 3, 160 & 0 & 97.5 & 56.2 & 0.963 & 0.993 (layer 25) & 27 \\
depth 2, lottery, $pr/b=0.95$ & 4, 160 & 0 & 97.9 & 62.1 & 0.968 & 0.993 (layer 25) & 27 \\
depth 2, checked, $pr/b=1.07$ & 1, 100 & 94 & 98.8 & 95.8 & 1.000 & 1.000 (layer 28) & 27 \\
depth 2, checked, $pr/b=1.07$ & 2, 100 & 96 & 99.2 & 97.1 & 1.000 & 1.000 (layer 33) & 27 \\
depth 2, lottery, $pr/b=1.07$ & 1, 100 & 87 & 99.2 & 95.8 & 0.999 & 0.999 (layer 27) & 27 \\
depth 2, checked, $pr/b=1.25$ & 1, 100 & 96 & 99.2 & 98.3 & 1.000 & 1.000 (layer 28) & 27 \\
depth 2, checked, $pr/b=1.25$ & 2, 100 & 98 & 99.2 & 98.8 & 1.000 & 1.000 (layer 35) & 27 \\
depth 2, checked, $pr/b=1.60$ & 1, 100 & 99 & 99.2 & 99.6 & 1.000 & 1.000 (layer 28) & 27 \\
depth 2, checked, $pr/b=2.00$ & 1, 100 & 99 & 99.2 & 99.2 & 1.000 & 1.000 (layer 28) & 27 \\
depth 2, checked, $pr/b=3.00$ & 1, 100 & 100 & 99.6 & 99.2 & 1.000 & 1.000 (layer 25) & 27 \\
depth 2, lottery, $pr/b=3.00$ & 1, 100 & 100 & 99.2 & 99.6 & 1.000 & 1.000 (layer 28) & 27 \\
depth 2, sharpening ($b=0$) & 1, 100 & 99 & 99.6 & 99.6 & 1.000 & 1.000 (layer 28) & 27 \\
\midrule
depth 2 sharpened, checked, $pr/b=0.60$ & 1, 100 & 0 & 99.2 & 50.4 & 0.975 & 0.999 (layer 25) & 27 \\
depth 2 sharpened, checked, $pr/b=0.90$ & 1, 100 & 49 & 99.2 & 90.0 & 1.000 & 1.000 (layer 28) & 27 \\
depth 2 sharpened, checked, $pr/b=1.10$ & 1, 100 & 95 & 99.2 & 96.7 & 1.000 & 1.000 (layer 28) & 27 \\
depth 2 sharpened, checked, $pr/b=1.60$ & 1, 100 & 99 & 98.8 & 99.2 & 1.000 & 1.000 (layer 25) & 27 \\
\midrule
depth 2, cut after lottery, $p_t\,r/b=0.93$ & 1, 100 & 83 & 99.6 & 94.2 & 1.000 & 1.000 (layer 28) & 27 \\
depth 2, cut after checked, $p_t\,r/b=1.06$ & 1, 100 & 99 & 99.6 & 99.6 & 1.000 & 1.000 (layer 25) & 27 \\
\midrule
depth 3, checked, $pr/b=0.50$ & 1, 100 & 0 & 89.2 & 50.8 & 0.895 & 0.934 (layer 27) & 29 \\
depth 3, checked, $pr/b=0.80$ & 1, 100 & 0 & 89.6 & 54.6 & 0.924 & 0.956 (layer 27) & 29 \\
depth 3, checked, $pr/b=0.92$ & 1, 160 & 20 & 90.4 & 65.8 & 0.953 & 0.958 (layer 28) & 29 \\
depth 3, checked, $pr/b=0.92$ & 2, 160 & 47 & 90.4 & 73.3 & 0.963 & 0.964 (layer 28) & 29 \\
depth 3, checked, $pr/b=0.92$ & 3, 160 & 36 & 87.9 & 67.1 & 0.957 & 0.962 (layer 28) & 29 \\
depth 3, lottery, $pr/b=0.92$ & 1, 160 & 0 & 89.2 & 52.1 & 0.935 & 0.955 (layer 27) & 29 \\
depth 3, lottery, $pr/b=0.92$ & 2, 160 & 0 & 88.3 & 52.9 & 0.928 & 0.956 (layer 27) & 29 \\
depth 3, lottery, $pr/b=0.92$ & 3, 160 & 0 & 89.6 & 54.6 & 0.937 & 0.958 (layer 27) & 29 \\
depth 3, checked, $pr/b=1.07$ & 1, 100 & 71 & 90.0 & 85.0 & 0.972 & 0.973 (layer 28) & 29 \\
depth 3, checked, $pr/b=1.07$ & 2, 100 & 77 & 89.6 & 90.0 & 0.971 & 0.972 (layer 31) & 29 \\
depth 3, checked, $pr/b=1.25$ & 1, 100 & 88 & 90.0 & 90.0 & 0.972 & 0.974 (layer 28) & 29 \\
depth 3, checked, $pr/b=2.00$ & 1, 100 & 93 & 89.2 & 92.1 & 0.972 & 0.973 (layer 28) & 29 \\
depth 3, sharpening ($b=0$) & 1, 100 & 93 & 88.8 & 89.6 & 0.969 & 0.970 (layer 28) & 29 \\
depth 3, second sharpening ($b=0$) & 1, 100 & 97 & 90.8 & 88.8 & 0.967 & 0.970 (layer 28) & 29 \\
\midrule
depth 3 sharpened, checked, $pr/b=0.93$ & 1, 100 & 38 & 91.2 & 78.8 & 0.971 & 0.973 (layer 31) & 29 \\
depth 3 sharpened, checked, $pr/b=1.07$ & 1, 160 & 78 & 91.2 & 89.6 & 0.976 & 0.976 (layer 28) & 29 \\
depth 3 sharpened, checked, $pr/b=1.07$ & 2, 160 & 82 & 90.0 & 90.8 & 0.972 & 0.975 (layer 28) & 29 \\
depth 3 sharpened, checked, $pr/b=1.07$ & 3, 160 & 82 & 91.2 & 90.4 & 0.981 & 0.983 (layer 28) & 29 \\
\bottomrule
\end{tabular}

\end{table}

\section{The race in detail}
\label{app:race}

\paragraph{The lottery does what it should.}
We took eleven matched pairs of runs, each pair a checked run and a lottery run, eight on the depth-2 base and three on the depth-3 base.
In every run we graded each UNDERIVABLE attempt on an underivable training prompt with the checker that pays the checked runs, rebuilding each prompt's instance from the RL data.
In the checked runs the regrade matches the recorded reward on all 178,121 attempts at depth 2 and all 33,478 at depth 3.
In the lottery runs the reward paid 77.2\% of valid attempts and 77.3\% of invalid ones at depth 2, against a target of 77.3\%, and 66.2\% and 65.7\% at depth 3, against a target of 66.2\%.
So the lottery ignores validity, as intended.

\paragraph{Validity in training.}
Here validity is the fraction of attempts in the training samples that the checker accepts.
\Cref{tab:lotteryskill} compares the checked and lottery runs at the last 10-step bin in which both runs of a pair still make at least 20 attempts.
At $pr/b=1.07$ and 3.0, where the lottery's attempts stay common to the end, the lottery's validity ends at 0.77 (0.71--0.79 throughout), while the checked runs reach 0.86 and 0.87.
In these two pairs the lottery runs make over 2,000 attempts in each 10-step bin.
Below the threshold, the lottery's attempts fall below about 100 per 10-step bin within 40 to 80 steps.
At depth 2 a trickle of attempts continues after that in two pairs, and at depth 3 all three lottery runs keep making a few attempts, up to 48 per bin, through step 160.
In eight of the nine pairs below the threshold, the lottery's last bin in the table holds 20 to 32 attempts (201 in the ninth).
These few attempts are the ones that survived selection.
In these bins the checked run's validity is higher in six of the nine pairs.
Over all eleven pairs, it is higher in eight, by 0.06 on average (95\% $t$ interval $-0.01$ to 0.13, one-sided sign test 0.11).
Pooled over the four seeds at $pr/b=0.95$, the checked runs' validity rises from 0.80 in steps 1--10 to 0.90 in steps 151--160.
Run by run, the checked runs' validity is 0.78--0.81 in steps 1--10 and 0.83--0.89 in the last bin of \cref{tab:lotteryskill}, and the four checked runs at $pr/b=0.95$ reach 0.89--0.91 by step 160.
The lottery's validity stays at 0.78--0.86 until its attempts run out, and its last 10-step bin with at least 100 attempts holds 113, from one seed.
On the depth-3 base at $pr/b=0.92$, the checked runs' validity rises from 0.66--0.67 in steps 1--10 to 0.81--0.85 in steps 151--160, so $p_t\,r/b$ in training goes from 0.92--0.93 to 1.12--1.18.

\paragraph{The rise is improvement.}
A rise in validity could also come from a model that attempts only the instances it already solves.
To separate the two, we sampled the base 16 times on each of 1,000 underivable training instances and graded the answers with the checker.
This gives the base's own validity on each of the 995 instances that drew at least one attempt (0.773 pooled).
For the runs at $pr/b=0.95$ we then took every attempt in the training samples on one of these instances and compared its validity with the base's validity on the same instance.
In the checked runs, pooled over the four seeds, the base's validity on the attempted instances stays within about 0.03 of its value on all instances in every 10-step bin, so the model keeps attempting instances of the usual difficulty.
Over the same bins, the validity of these attempts rises from 0.79 in steps 1--10 to 0.90 in steps 151--160, against 0.78 for the base on the same instances and 0.77 on all of them.
So of the 12-point gap between these attempts and the base on all instances, selection accounts for about 1 point and improvement for about 11.
Unrounded, the checked runs' attempts in steps 151--160 are valid 0.898 of the time, against 0.784 for the base on the same instances and 0.773 on all instances.
Under the lottery, while attempts are common (steps 1--40), they are within 0.01 of the base's validity on the same instances.
Once they become rare (steps 61--80), the few that remain are on easier instances, where the base was valid 0.85--0.95 of the time, and the checker accepts 0.84--0.87 of them, no more than the base on the same instances.

\begin{table}[htbp]
\caption{Validity of the attempts in the eleven matched pairs of checked and lottery runs, at the last 10-step bin in which both runs make at least 20 attempts, on the depth-2 base (top) and the depth-3 base (bottom).
Each entry gives the validity, with the number of attempts in parentheses, and the difference is checked minus lottery, computed before rounding.}
\label{tab:lotteryskill}
\centering\scriptsize
\setlength{\tabcolsep}{4pt}
\begin{tabular}{lccccc}
\toprule
$pr/b$ & seed & steps & checked & lottery & difference \\
\midrule
0.80 & 1 & 31--40 & 0.88 (108) & 0.70 (20) & $+0.18$ \\
0.93 & 1 & 91--100 & 0.85 (1814) & 0.90 (20) & $-0.05$ \\
0.95 & 1 & 51--60 & 0.85 (845) & 0.96 (24) & $-0.11$ \\
0.95 & 2 & 151--160 & 0.89 (1695) & 0.64 (22) & $+0.25$ \\
0.95 & 3 & 61--70 & 0.86 (1454) & 0.81 (32) & $+0.05$ \\
0.95 & 4 & 41--50 & 0.83 (1073) & 0.81 (201) & $+0.02$ \\
1.07 & 1 & 91--100 & 0.86 (2349) & 0.77 (2334) & $+0.09$ \\
3.0 & 1 & 91--100 & 0.87 (2474) & 0.77 (2474) & $+0.10$ \\
\midrule
0.92 & 1 & 51--60 & 0.82 (397) & 0.77 (26) & $+0.05$ \\
0.92 & 2 & 101--110 & 0.75 (1021) & 0.66 (29) & $+0.09$ \\
0.92 & 3 & 61--70 & 0.81 (160) & 0.85 (26) & $-0.03$ \\
\bottomrule
\end{tabular}
\end{table}

\section{A model of the race}
\label{app:racemodel}

\paragraph{The model.}
The model follows \cref{prop:step} through training, with one assumption added about how the checker raises the validity of attempts.
Each underivable instance $i$ has verdict log-odds $f_i+X(t)$, and each derivable instance $j$ has log-odds $f_j+Y(t)$.
The offsets $f_i$ and $f_j$ are the base's teacher-forced log-odds on the 1,000 underivable and 1,000 derivable training instances of \cref{app:shared}, so the spread of the base's certainty is measured.
Instance $i$'s attempts are valid with probability $p_i(t)=p_i+(1-p_i)\,g(t)$, where $p_i$ is the base's validity on it (16 samples each, \cref{app:std}) and $g(t)$ is a skill gain shared by all instances.
Each step samples a batch of 31 underivable and 33 derivable prompts with 8 answers each, and pays them as in training.
The summed group pushes of the two classes, $P_t$ and $P_o$ (where $t$ marks the class at stake, as in \cref{prop:step}), pass through Adam's momentum ($\beta_1=0.9$) and then move the offsets by $\Delta X\propto P_t+sP_o$ and $\Delta Y\propto sP_t+(s/s')P_o$, with the couplings $s=1.03$ and $s'=0.86$ measured on the base.
We use two versions of this update.
In the linear version, the offsets move by a constant $\alpha$ times the momentum, and in the Adam-normalized version, by $\alpha$ times the momentum divided by the square root of Adam's second moment, as if the verdict's direction dominated it.
Under the checker the skill gain grows by $\beta\,A(t)\,(g_{\max}-g)$ per step, where $A(t)$ is the share of answers on underivable prompts that are attempts, and under the lottery it stays at 0.
A run wins if its greedy honesty, the share of underivable instances with $f_i+X>0$, is at least 60\% at its last evaluation.

\paragraph{The fit.}
None of the constants is fit on the runs that test the race, the depth-2 runs at $pr/b=0.80$, 0.93, and 0.95.
The skill constants, $\beta=0.011$ and $g_{\max}=0.63$, are fit to the difference in validity between the checked and lottery runs of seed 1 above the threshold, which see the same prompts in the same order, and to the rate $p_t$ that the checked runs reach after 100 steps.
The step scale $\alpha$ is fit, for each version of the update, to the honest share in the training samples of the checked runs at $pr/b=0.40$, 0.60, 1.07, 1.25, 1.6, 2.0, and 3.0 and the lottery runs at 1.07 and 3.0.
The two versions fit these runs equally well (root-mean-square error 0.037 and 0.039).
Both reproduce every fitting run's outcome.
The linear version's dips above the threshold are too shallow, and its collapse at $pr/b=0.40$ is too fast (45\% honest in the training samples at step 20, against 77\%).
The Adam-normalized version starts the collapses at $pr/b=0.40$ and 0.60 too slowly, by several points in the first steps.

\begin{figure}[htbp]
\centering
\includegraphics[width=\textwidth]{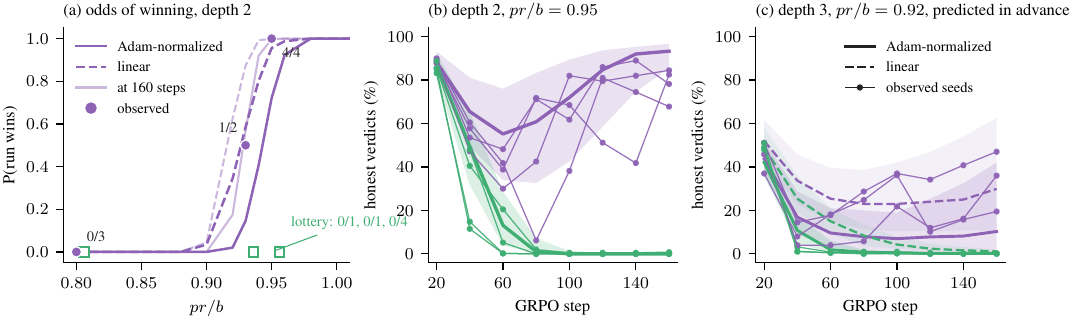}
\caption{The model of the race, fit without the runs it predicts, against the race runs on the depth-2 base (a, b) and the depth-3 base (c).
(a) The predicted chance that a checked run from the base wins, against $pr/b$, for the two versions of the update after 100 steps and for each after 160 steps, with the observed outcomes of the checked runs (circles) and lottery runs (squares).
(b) At $pr/b=0.95$, the median and 10--90\% band of honest verdicts over simulated runs of the Adam-normalized version, for the checker (purple) and the lottery (green), with the observed seeds.
(c) The depth-3 base at $pr/b=0.92$, with the medians and 10--90\% bands of both versions (Adam-normalized solid, linear dashed), computed from the depth-3 fit before these runs were done, and the observed seeds.}
\label{fig:racemodel}
\end{figure}

\paragraph{Cold predictions.}
We simulated 500 runs at each ratio, with sampling noise only.
\Cref{fig:racemodel}(a) shows the predicted chance of winning against $pr/b$, with the race runs' outcomes.
Every checked run at 0.80 and every lottery run below the threshold is predicted to collapse, as they do.
At 0.93 the chance of winning after 100 steps is 0.59 in the linear version and 0.15 in the Adam-normalized one, and one of two seeds won.
At 0.95 it is 1.00 after 160 steps in both, and all four seeds won.
In the Adam-normalized version the predicted median at 0.95 dips and recovers as the observed seeds do (\cref{fig:racemodel}(b)), but the observed dips are deeper and last longer.
From step 40 on, at least one seed lies below the predicted 10--90\% band at every evaluation, as can be expected of a band that holds only sampling noise, and the predicted final level, 93\%, is above the observed 68--85\%.
In the simulations, as in the runs, the outcome near the edge is settled only after the first steps.
In the runs, the two checked seeds at $pr/b=0.93$ are as honest as each other, and their attempts as valid, through step 30.
The seed that later wins then falls faster (52\% against 78\% honest in the training samples of steps 31--40), and the two cross only around steps 55--65, when it turns up and the other keeps falling.

\paragraph{Stability.}
To see how much the predictions lean on any one fitting run, we left out in turn each of the 11 runs behind the step scale and each of the 7 runs behind the skill law, and refit the constants.
For each refit we simulated 150 runs per ratio.
The predicted edge, the ratio at which the chance of winning after 100 steps reaches one half, stays between $pr/b=0.918$ and 0.932 in the linear version (0.926 with every run) and between 0.940 and 0.945 in the Adam-normalized one (0.943).
The chance of winning at 0.93 after 100 steps stays between 0.46 and 0.69 in the linear version and between 0.11 and 0.16 in the other, and at 0.95 after 160 steps it is 0.98 or more in both.
So the predictions depend much more on the choice between the two versions than on any single fitting run.

\paragraph{Depth 3.}
For the depth-3 base we fit the same constants on depth-3 runs only (at $pr/b=0.50$, 0.80, 1.25, and 2.0), with its own couplings ($s=0.98$, $s'=0.95$), and with per-instance validities read off 101 prompts in the first three steps of these runs and rescaled to the base's $p$.
The fits are rougher than at depth 2.
We recorded the prediction in this paper's version history before the runs at $pr/b=0.92$ were done.
Each of three checked seeds, run for 160 steps, would win with probability 0.13 in the linear version and 0.04 in the Adam-normalized one, so most likely none would win and at most one would.
Each of three lottery seeds would collapse.
The depth-3 base's edge is predicted at about $pr/b=0.95$.

\paragraph{The depth-3 runs.}
No checked seed won within 160 steps, and every lottery seed collapsed, as predicted.
One checked seed ends collapsed, at 19.5\%, and two are undecided, while the predicted chance that a seed collapses was 0.35 in the linear version and 0.73 in the Adam-normalized one.
By step 40 the checked seeds fall to 4--15\%, below the linear version's 10--90\% band (23--46\%), and two of the three fall below the Adam-normalized one's (9--29\%).
They then climb back faster than the Adam-normalized version predicts, and at step 160 all three lie inside the linear version's band (11--63\%), while one lies above the Adam-normalized band (1--42\%) (\cref{fig:racemodel}(c)).
The lottery seeds collapse faster than either version predicts, to 1--3\% at step 40 against predicted medians of 25\% and 11\%.
From step 60 on, every checked seed is more honest than every lottery seed, as the predicted medians are.
Both checked seeds at $pr/b=1.07$, which the model gives a chance of winning of 1 in both versions, survive.

\paragraph{What the model leaves out.}
Its only noise is that of sampling a batch, so its spread of outcomes is a lower bound on the real one.
All instances of a class move by a common shift of their log-odds.
The skill law, proportional to the attempts the checker grades and saturating, is an assumption fit to ten noisy bins.

\section{Finite time}
\label{app:finite}

\paragraph{$\kappa$ on Datalog.}
Consider the five runs from the depth-2 base above the threshold ($pr/b=1.07$ and 1.25 with two seeds each, and 1.6).
For these runs we computed $W_t$, $W_o$, $c$, and $\tilde p$ at every step from the training samples, which have 64 prompts and 8 answers per step.
We used the unbiased estimate of $h(1-h)$ from 8 samples and pooled over 7-step windows.
The ratio $W_o c/(W_t b)$ is mostly between 1 and 4 over the first eight steps, and below 0.2 from steps 13--18 until at least step 35, because by then almost no derivable prompt draws both verdicts in its group.
Later in the runs it can rise again, because $W_t$ also becomes small once nearly every underivable prompt is answered honestly.
Over the first 35 steps the weighted ratio $\tilde p\,r/b$ stays between about 0.9 and 1.3 at $pr/b=1.07$, and is higher at 1.25 and 1.6.
With the measured coupling $s=1.03$, $1+\kappa$ starts above $\tilde p\,r/b$ in all five runs and first falls to it at steps 13, 16, 15, 11, and 18.
For any $s$ between 0.8 and 1.1 these steps move by at most one.
In the run at $pr/b=1.07$ of \cref{fig:dip}(a), the two sides then stay within about 0.1 of each other until step 28.
Over this run the margin $\tilde p\,r/b-1-\kappa$ averages $-2.4$ over steps 1--12, $+0.01$ over steps 13--28, and $+0.13$ over steps 29--35, so the steps between 13 and 28 push little either way.
Honesty in the training samples bottoms out later, at steps 29--34.
If we sum the push of \cref{prop:step} step by step, using each batch's realized group pushes, the predicted turn comes at steps 13--26.
If we first average the push as Adam's momentum does ($\beta_1=0.9$), it comes at steps 30, 23, 23, 22, and 23, and honesty bottoms out at 34, 29, 31, 29, and 31.
Momentum thus delays the predicted turn by 4 to 9 steps, to steps 22--30, which is 4 to 8 steps before honesty bottoms out.
We resampled the prompts within each step (both classes) and recomputed both the predicted turn and the step at which honesty bottoms out.
The 90\% interval of the gap between them includes 0 in four of the five runs, and its lower end is 0 in the fifth (the run at $pr/b=1.25$, seed 1).
We also split the underivable prompts by whether their own success rate is below $b/r$ (14--21\% of them are).
The two groups fall and turn together, so the lag holds within each group of prompts.
The training samples of the checked runs at $pr/b=1.6$--3.0 also dip, to 92\%, 94\%, and 96\% honest around step 31, although these runs stay 97--100\% honest at every evaluation.

\section{Generous, then cut}
\label{app:cut}

\paragraph{Bluffs at the lower ratio.}
After 100 generous steps under the checker the model is 99.5\% honest, and after the cut it samples bluffs again.
On underivable training prompts after the cut, the share of bluffs is 0\% in steps 1--10, 0.8\% in steps 11--20, 8.3\% and 8.0\% in steps 21--40 and 41--60, and 3.5\% and 1.8\% in steps 61--80 and 81--100.
In all, 99 groups in steps 21--100 hold both a bluff and an attempt.
In each 20-step bin, 81--94\% of these groups favor the attempt, which is valid 0.87--0.92 of the time, against $b/r=0.81$.
So the bluffs come back and lose.
When the generous phase is run under the lottery instead, the share of bluffs after the cut climbs to 25\% in steps 41--60 and 57\% in steps 61--80.

\begin{figure}[htbp]
\centering
\includegraphics{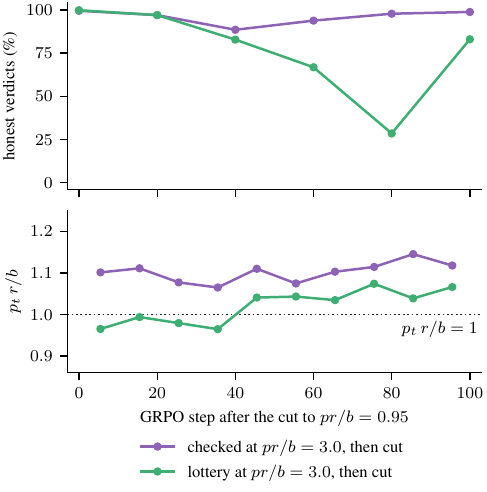}
\caption{A generous phase with and without a checker.
The depth-2 base is trained for 100 steps at $pr/b=3.0$ under the checker (purple) or under a lottery with the same expected payoff (green), and then cut to $pr/b=0.95$ under the checker.
Top: honest verdicts at each evaluation.
Bottom: $p_t\,r/b$ as in \cref{fig:cut}.}
\label{fig:cutlot}
\end{figure}

\paragraph{From a lottery-trained model.}
Consider the depth-2 base trained for 100 steps at $pr/b=3.0$ under the lottery.
It stays honest (98--100\%), and its success rate stays flat, ending at $p_t=0.76$, against 0.87 under the checker.
We then cut its ratio to $pr/b=0.95$, with $p$ the base's rate, and train it under the checker.
It starts at $p_t\,r/b=0.93$, falls to 29\% honest at step 80, and ends 83\% honest at step 100, by which point $p_t$ has risen to 0.85.
This run starts below $p_t\,r/b=1$ and still ends honest, so the criterion of step 2 of the recipe may be conservative.
The run cut from the model trained under the checker stays 89--99\% honest (\cref{fig:cutlot}).
Both end honest, and the check in the generous phase avoids the dip.

\section{Further related work}
\label{app:related}

Work posted in October and November 2026 \citep{symce2026,zhan2026precisecoder,deng2026verdict,arora2026graceful,li2026multireward,chen2026safety,mao2026trophies,li2026audit,he2026rlec,ding2026collective,wang2026forecasting,fel2026shaping} is concurrent with ours.

\paragraph{Thresholds and collapse laws.}
\citet{zhang2025lead} build a wrong-answer penalty whose break-even is a fixed confidence near one half.
\citet{moya2026verifier} give conditions under which a fixed imperfect verifier raises reward while correctness falls.
Their conditions are stated on the rates of change of acceptance and of the hacked share.
\citet{peng2025unsolvable} derive when declaring a problem unsolvable beats attempting it.
They add a false-rejection penalty so that training does not collapse into universal rejection.
\citet{che2026abstention} derive a collapse toward refusing everything under error penalties when answering everything loses score in expectation, and observe that it silences questions the model still solves.
\citet{egashira2026delay} find that systematic false positives can collapse training in ways that aggregate rates do not predict.
Here $b$ and $p$ do predict it.
After anti-sycophancy RL that rewards correct true-or-false verdicts, the share of true statements that a model calls false rises from 38.5\% to 51.5\% \citep{cheng2026euston}, much as our code models call correct programs buggy at high payoff ratios (\cref{sec:toomuch}).
RL for debugging that is rewarded by unit tests alone teaches models to hallucinate bugs in code that is already correct \citep{zhan2026precisecoder}.
A code judge rewarded for its verdict alone favors the majority verdict without testing the candidate, and for this reason \citet{deng2026verdict} pay little for an unaccepted verdict without a counterexample that survives execution.
In visual grounding, refusal training trades accuracy on the positive class for refusals \citep{yang2026refusal}.
In GRPO safety training, refusals learned on harmful prompts can spill over into refusals of benign ones \citep{chen2026safety}.

\paragraph{Schedules.}
A shaping reward that is not potential-based can change the optimal policy \citep{ng1999policy}, which is why the generous phase has to end.

\paragraph{Training dynamics.}
Correct actions that GRPO does not sample can lose probability mass even as total correct mass grows \citep{plyusov2026fgrpo}.
A prompt with zero success probability stays at zero \citep{mroueh2025effective}.
This is the absorbing branch of the race, in which the attempts die out and the collapse is permanent.
An action whose payoff rises only when it is taken, and by less each time, is an arm of a rested rising bandit \citep{metelli2022rising}.
The honest action under the checker is such an action, and the lottery makes it a stationary one.
In the theory of \citet{huang2026curriculum}, skill grows on problems the model already partly solves and spreads from there to slightly harder ones.
\citet{wu2026rebound} describe coding RL runs that read as a race running the other way, in which a temporary retreat to honest solving comes before a hack takes over when honest solving earns too little reward.
Random rewards can improve some models, which \citet{shao2025spurious} attribute to a clipping bias.
Whether random rewards help depends on the checkpoint, even on synthetic problems \citep{mao2026trophies}.
On a contamination-free arithmetic task, the gains are marginal and unstable \citep{wu2025memorization}.
Base success alone does not determine what RLVR does to a prompt \citep{yang2026kernel}.
Our threshold predicts the side a run ends on from $p$ and the competing payoff together.
How an update on one example moves the others has been studied for fine-tuning \citep{ren2025learning}.
There, refusal spills over onto benign prompts that are close to toxic ones in this kernel (the inner product of gradients) \citep{lu2026discern}.
Within a single prompt, preference training can move probability onto responses of the opposite meaning, driven by preferred and dispreferred responses with similar hidden embeddings \citep{razin2025unintentional}.
How an update on one prompt moves the others under RL post-training is studied by \citet{barakat2026interference}.
In label-free RLVR, each problem sharpens more slowly as the number of training problems grows, while a direction that problems share through their parameters grows at a rate independent of that number, which explains why training collapses only when it uses many problems \citep{ding2026collective}.
Our coupling $s$ measures such sharing between the two classes.
First-order forecasts fixed before training also predict how RLHF responds to edits of the reward model, and their ranking of controlled edits holds over the first 30 steps but reverses by step 200, while real reward-model upgrades keep their order \citep{wang2026forecasting}.
Our step condition is a first-order forecast of this kind, and it gives only the direction of a step.
With aligned features, the representation part of a rewarded update raises the sampled token's probability in every context \citep{tomihari2026dynamics}.

\paragraph{What RL changes.}
\citet{shenfeld2025razor} argue that online RL forgets less than SFT because it stays close to the base in KL divergence.
\citet{shi2026why} find that RL preserves base features where SFT adds new ones.
That a probe survives an RL collapse fits this picture.

\paragraph{Reward hacking, abstention, and refutation.}
Adjacent topics include detecting reward hacking from representations \citep{wu2026rebound,bergen2026monitoring}, the decoupling of reasoning from answers \citep{maniyar2026decoupling}, abstention as an RLVR objective \citep{zhai2026abstain}, and removing sandbagging \citep{ryd2026sandbagging}.
\citet{szablewski2026lottery} warn that interpretability findings on model organisms depend on how the organisms are trained.
This is why we use several bases, two model families, and three tasks.
Feedback from an evaluator who cannot fully check a claim can teach a policy to mislead \citep{lang2024deceive,wen2024mislead}.
Coding agents given impossible tasks often exploit the tests \citep{zhong2025impossiblebench}.
In our setting the unverifiable claim is part of the reward, and the honest alternative has a measured success rate.
Reward misspecification can produce phase transitions in which a more capable policy exploits the reward more \citep{pan2022effects}.
Here the dependence runs the other way, since a model that is better at producing certificates bluffs less.
Training a model to give answers that a verifier can check is the setting of prover-verifier games \citep{anil2021learning,kirchner2024prover}, whose verifiers are trained against ground-truth labels.
Our reward instead pays DERIVABLE (on code, CORRECT) without a certificate, whether or not it is true.
Finding counterexamples can be harder for models than solving the problem \citep{sinha2025falsify}.
\citet{li2026disprove} train this skill with a counterexample reward verified in Lean, which is the closest certificate-graded refutation reward we know of.
The differential oracle of our code task, which compares the program under test with the reference on one call, is the reward, or a term of the reward, in RL for test generation \citep{liu2025harnessllm,xu2026adversarial,liang2026oracle}.
That setting has no option of calling a program correct.
In logical reasoning, \citet{gangwani2026logicverified} see recognition of the unknown class fall slightly under GRPO with an outcome reward.
\citet{liu2025anchor} find that GRPO with an outcome reward collapses to always answering Unknown on one of two deduction tasks and learns the other only partly.

\section{Code and data release}
\label{app:tables}

We will release the code and data with a future version of this paper.
The release will contain the Datalog instance generator and checker, the reward tables, the SFT and GRPO configurations, the probe scripts, the code task, and every surviving result file.
It will also contain an inventory that maps each number in this paper to the file it was read from, together with the scripts that draw each figure and table.

\end{document}